\documentclass[10pt,twocolumn,letterpaper]{article}
\usepackage{iccv}
\usepackage{times}
\usepackage{epsfig}
\usepackage{graphicx}
\usepackage{amsmath}
\usepackage{amssymb}
\usepackage{xcolor}
\usepackage{booktabs}
\newcommand{\tabincell}[2]{\begin{tabular}{@{}#1@{}}#2\end{tabular}}
\usepackage{multirow}
\usepackage{booktabs}

\usepackage{subfigure}
\usepackage{pifont}
\newcommand{\cmark}{\ding{51}}%
\newcommand{\xmark}{\ding{55}}%
\newcommand*{\affaddr}[1]{#1} 

\newcommand*{\email}[1]{\texttt{#1}}
\usepackage[toc,page]{appendix}
\usepackage[pagebackref=true,breaklinks=true,letterpaper=true,colorlinks,bookmarks=false]{hyperref}

\iccvfinalcopy 

\ificcvfinal\fi

\begin{document}
	
	\title{Grab What You Need: Rethinking Complex Table Structure Recognition with Flexible Components Deliberation}
	
\author{
	Hao Liu\textsuperscript{\dag\thanks{Equal contribution. \endgraf \textsuperscript{\dag}Corresponding author, e-mail: \email{\textbf{ivanhliu@tencent.com}}.}} \quad Xin Li\textsuperscript{*} \quad Mingming Gong \quad Bing Liu \quad \\ Yunfei Wu \quad Deqiang Jiang \quad Yinsong Liu \quad Xing Sun\\		
	\affaddr{Tencent YouTu Lab} 
}
\maketitle
	
	\maketitle
	\ificcvfinal\thispagestyle{plain}\fi

	\begin{abstract}
		Recently, Table Structure Recognition (TSR) task, aiming at identifying table structure into machine readable formats, has received increasing interest in the community. While impressive success, most single table component-based methods can not perform well on unregularized table cases distracted by not only complicated inner structure but also exterior capture distortion. In this paper, we raise it as Complex TSR problem, where the performance degeneration of existing methods is attributable to their inefficient component usage and redundant post-processing. To mitigate it, we shift our perspective from table component extraction towards the efficient multiple components leverage, which awaits further exploration in the field. Specifically, we propose a seminal method, termed GrabTab, equipped with newly proposed Component Deliberator. Thanks to its progressive deliberation mechanism, our GrabTab can flexibly accommodate to most complex tables with reasonable components selected but without complicated post-processing involved. Quantitative experimental results on public benchmarks demonstrate that our method significantly outperforms the state-of-the-arts, especially under more challenging scenes.
	\end{abstract}
	
	\section{Introduction}
		\begin{figure}[htb]
		\begin{center}
			\includegraphics[width=1\linewidth,height=0.8\linewidth]{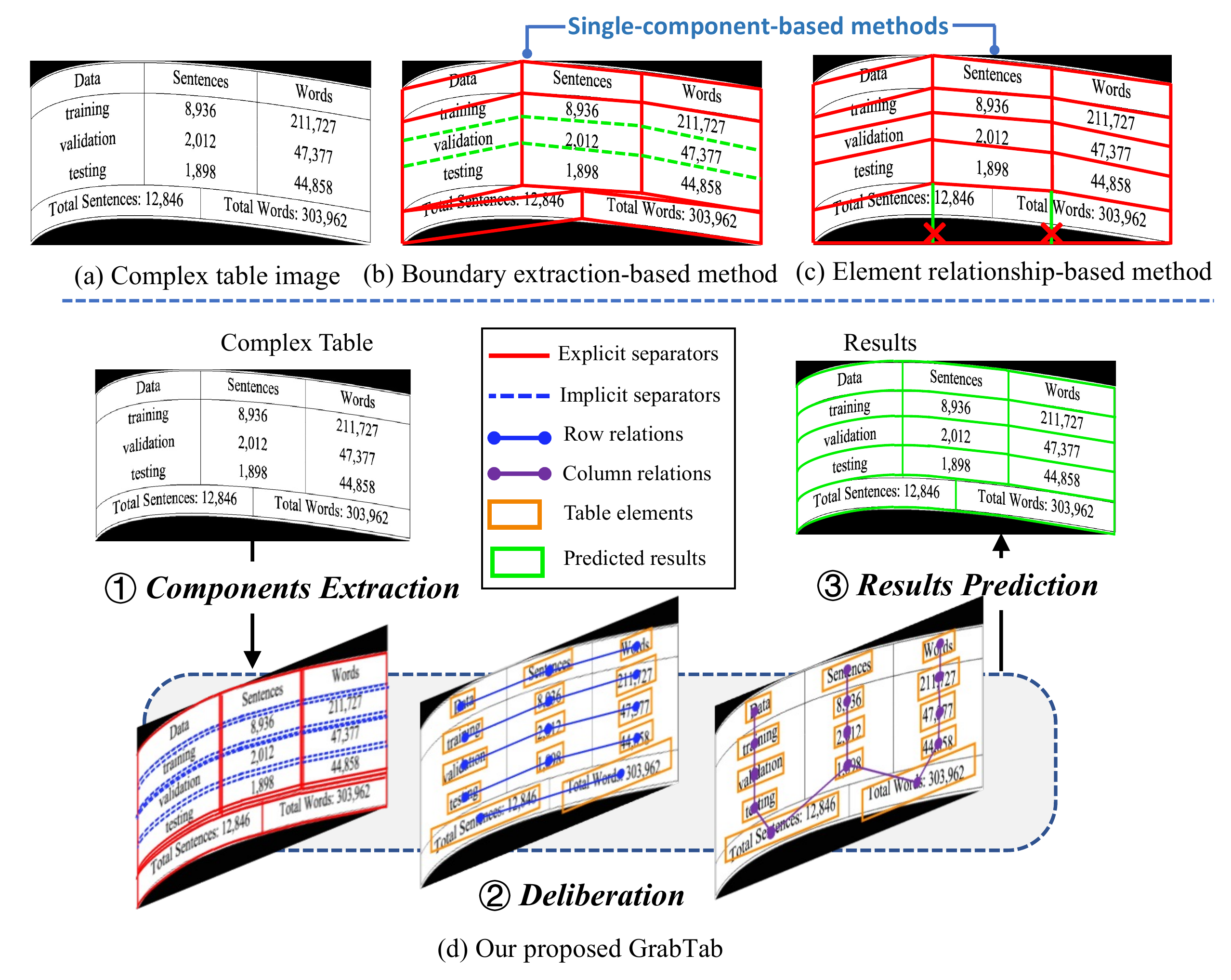}
			\\
		\end{center}
		\caption{Illustration of motivation of the proposed GrabTab. (a)~Complex table image. (b)~Boundary extraction-based methods. (c)~Element relationship-based methods. As merely one single table component leveraged, the predicted cell boundaries could suffer from ``boundary missing''~(green dashed lines) or ``over-prediction''~(green solid lines) problem. (d)~Our proposed GrabTab. A set of table components, including row/column relations, table elements, visual explicit and implicit separators, are ``deliberated'' by our GrabTab, where informative clues are flexibly picked up and assembled, which is more versatile for various complex table layouts. Best viewed in color.
		}\label{fig:moti}
	\end{figure}
	With the fast-paced development of digital transformation, Table Structure Recognition~(TSR) task, aiming at parsing table structure from a table image into machine-interpretable formats, which are often presented by both table cell physical coordinates ~\cite{schreiber2017deepdesrt,paliwal2019tablenet,khan2019table,tensmeyer2019deep,chi2019complicated,rethinkingGraphs,raja2020table,zheng2021global,qiao2021lgpma,liu2021show,long2021parsing} and their logical relationships~\cite{li2020tablebank,end2end1}. It has received increasing research interest due to the vital role in many document understanding applications~\cite{jauhar2016tables, li2016deep}. To date, several pioneers works~\cite{li2020tablebank,end2end1,schreiber2017deepdesrt,khan2019table,tensmeyer2019deep,chi2019complicated,rethinkingGraphs,raja2020table,zheng2021global,qiao2021lgpma,liu2021show,long2021parsing} have achieved significant progress in the filed, which can be mainly categorized into  \textit{boundary extraction-based methods}~\cite{schreiber2017deepdesrt, paliwal2019tablenet, khan2019table, tensmeyer2019deep,long2021parsing} and \textit{element relationship-based methods}~\cite{liu2021show, chi2019complicated,rethinkingGraphs,raja2020table} according to the component type leveraged.

	Unfortunately, the above single-component-based techniques can only yield promising results on regularized table cases, but struggle when processing more complicated cases as illustrated in Fig.~\ref{fig:moti}(b) and (c). However, with the popularization of mobile capture devices, the requirement on recognizing camera-captured tables has become increasingly imperative. In this scene, other than complicated table inner structure, geometrical distortion incurred by the capturing process becomes another distracting factor. In this paper, we define this more challenging TSR as \textit{Complex TSR task}. We attribute the performance degeneration of previous methods to their inefficient component usage and heavy dependence on rule-based post-processing. To be specific, given a complex table~(Fig.~\ref{fig:moti}(a)), boundary-based methods can better handle the visible boundary cells by directly predicting them, but suffer from ``boundary missing'' problem for those without explicit separations~(green dashed lines in Fig.~\ref{fig:moti}(b)). Comparatively, the relationship-based alternatives can overcome this issue by inferring cell boundaries from the element relationships, whereas the ``over-prediction'' of boundaries~(green solid lines in Fig.~\ref{fig:moti}(c)) is witnessed as the visible cell boundary clues are totally abandoned. To relieve these shortcomings, both methods resort to well-designed post-processing rules, however, they would become uncontrollable when distortion happens. This \textit{status quo} begs for a question: \textit{``Is there a versatile TSR solution to leverage merits of multiple components rather than merely single one, in the light of different complex table cases?''}. A straightforward way is to directly combine multiple components. Nevertheless, it is infeasible to implement as either component in the aforementioned methods is strongly coupled with corresponding post-processing rules, which could be mutually exclusive. For the Complex TSR, a few recent researches~\cite{long2021parsing, liu2022neural, lin2022tsrformer} have made attempts, whereas they only take further steps on more robust components~(relationships or boundaries) extraction while the deployment of multiple components is still rarely explored. 
	
	In human cognitive system~\cite{clancey2002simulating,anderson2005cognitive,olshausen1993neurobiological}, ``deliberation'' is one of common behaviors when human processing daily works, such as reading or analyzing table image. Specifically, a set of evident visual clues are perceived at a rough level and the final results are yielded by complementing them with implicit but necessary information after deliberation in a progressive way. Inspired by it, we in this paper introduce the ``deliberation'' mechanism and propose a novel method, termed \textbf{GrabTab}, tailored for Complex TSR problem, which can flexibly \textbf{Grabs} the needed information from a set of \textbf{Tab}le components and progressively assembles them to the final results, as demonstrated in Fig.~\ref{fig:moti}(d). 
	
	Concretely, we first go beyond canonical straight line-based method and propose a new table Separator Perceiver~(SP) based on Bézier curve, which can yield high-quality explicit~(solid red lines) and implicit separator~(dashed blue lines) proposals. Treating both types of separator proposals as hints, the newly designed Components Correlator~(CC) ``grabs'' useful information from table elements and their global relationships aggregated on the separator proposals shown in Fig.~\ref{fig:moti}(d). Afterward, requiring no sophisticated post-processing, our GrabTab directly predicts the logical index and ``grabs'' refined separators to constitute the final results through Structure Composer~(SC). Serving as core submodules, SP, CC and SC comprise our Components Deliberator~(CD) implementing progressive deliberation mechanism. Thanks to this mechanism and removal of complicated heuristic-based post-processing, our GrabTab exhibits prominent versatility, which can flexibly accommodate to most complex tables with reasonable components selected. Benefiting from the tailored design, our GrabTab method can achieve better performance compared to other TSR methods, especially for the complex table scenarios, as vividly validated by extensive experimental results. Conclusively, our contributions are summarized as:
	
	\begin{itemize}
		\item We reinspect the TSR task from the perspective of the efficient multiple components leverage, rather than single component extraction widely adopted by previous methods. To our best knowledge, we are the first to introduce deliberation mechanism and investigate its working patterns on component interaction for predicting complex table structure.
		
		\item We coin a novel and versatile method, GrabTab, tailored for Complex TSR problem, which is equipped with Components Deliberator consisting of Separator Perceiver, Component Correlator and Structure Composer, responsible for generation of high-quality separator proposals, multiple components correlation and composing refined separator into final results.
		
		\item Quantitative experimental results on public benchmarks demonstrate that our method can fully leverage components reciprocity for diversified complex table cases, without introducing extra complicated processes. Consequently, significant performance improvement is witnessed, especially under more challenging scenes.
		
	\end{itemize}

	\begin{figure*}[htb!]
		\centering
		\includegraphics[width=0.96\linewidth,height = 0.3\linewidth]{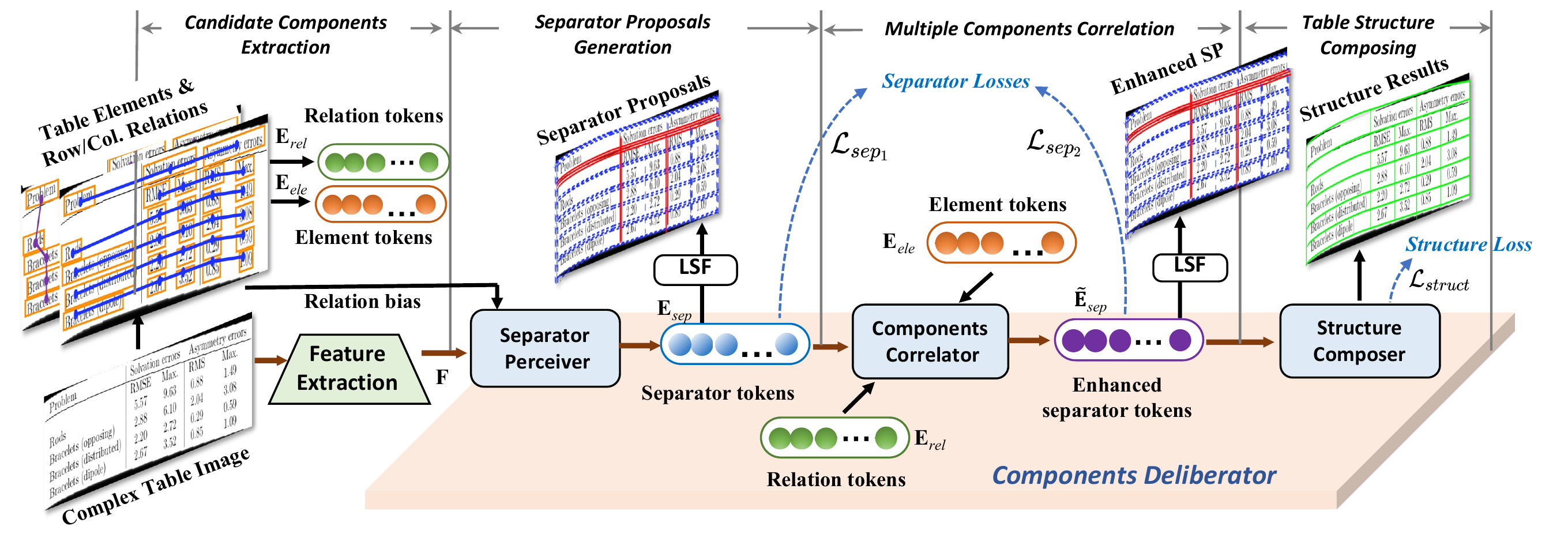}
		\vspace{-3mm}
		\caption{The architecture of our proposed GrabTab. Best viewed in color and zoomed in.} 
		\label{fig:arch}
		\vspace{-5mm}
	\end{figure*}

	\section{Related Work}
	In the early days, the table structure recognition~(TSR) field was once dominated by the traditional methods~\cite{green1995recognition, hirayama1995method, itonori1993table, kieninger1998t, tupaj1996extracting}. With the rise of deep learning, this field has witnessed a great change, bringing about remarkable performance improvement. In terms of the characteristics, they can be summarized into two categories. The first group are the single component-based methods, which rely on either boundary extraction~\cite{khan2019table, long2021parsing, paliwal2019tablenet, schreiber2017deepdesrt, tensmeyer2019deep} or element relationship~\cite{chi2019complicated, liu2021show, rethinkingGraphs, raja2020table}. And the second group are the markup generation models~\cite{li2020tablebank, nassar2022tableformer, end2end1}.
	\subsection{Single Component-based Methods}
	\noindent\textbf{Boundary extraction-based methods.}\quad
	In view of visual saliency, DeepDeSRT~\cite{schreiber2017deepdesrt} and TableNet~\cite{paliwal2019tablenet} are two representative works exploiting semantic segmentation to obtain table cell boundaries. With the same goal, bidirectional GRUs~\cite{khan2019table} extracts the boundaries of rows and columns in a context-driven manner. These methods, however, are at a loss when identifying cells that span over multiple rows and columns. The SPLERGE~\cite{tensmeyer2019deep} divides table into grid elements and merges adjacent ones to restore spanning cells, where the boundary ambiguity issue still remains unsolved. To attack it, a hierarchical GTE~\cite{zheng2021global} employs clustering algorithm while Cycle-CenterNet~\cite{long2021parsing} designs a cycle-pairing module that simultaneously detect table cells and group them into structured tables. Moreover, LGPMA~\cite{qiao2021lgpma} applies a soft pyramid mask when learning both global and local feature maps. However, in complex scenarios, satisfactory performance cannot be achieved through the subsequent heuristic structure recovery pipeline.
	
	With the flourish of transformer~\cite{vaswani2017attention}, several DETR~\cite{carion2020end}-like TSR methods are proposed to extract boundary in object detection paradigm. In work~\cite{smock2022pubtables}, a set of table elements are detected by DETR to model the hierarchical structure. TRUST~\cite{guo2022trust} has the similar spirit with SPLERGE~\cite{tensmeyer2019deep}, where the features of row/column separators are extracted for a further vertex-based merging module to predict the linking relations between adjacent cells, which thus still suffers from boundary ambiguity problem. TSRFormer~\cite{lin2022tsrformer} proposes a two-stage DETR framework that introduces a prior-enhanced matching strategy to solve the slow convergence issue of DETR and a cross attention module to improve localization accuracy of  row/column separators.
	
	\noindent\textbf{Element relationship-based methods.}\quad
	GraphTSR~\cite{chi2019complicated} enables information exchange between edges and vertices with the help of the graph attention blocks. The method~\cite{rethinkingGraphs} employs DGCNN~\cite{wang2019dynamic} to model interactions between visual and geometric feature vertices for their cell/row/column relations prediction. Also based on DGCNN, TabStruct-Net~\cite{raja2020table} is proposed for end-to-end training cell detection and structure recognition in a joint manner. FLAG-Net~\cite{liu2021show} is an another end-to-end method, which learns to reason table element relationships by modulating the aggregation of dense and sparse contexts in an adaptive way. Besides, CTUNet~\cite{li2022end} adopts the graph attention network with masked self-attention to effectively aggregate table cell nodes. To mine the intra-inter interactions between multiple modalities, NCGM~\cite{liu2022neural} designs the stacked collaborative blocks to model them in a hierarchical way. DocReL~\cite{li2022relational} uses Relational Consistency Modeling to excavate contextual knowledge more fully, so as to better learn relational representations in both local and global perspective.
	
	\subsection{Markup Generation-based Methods}
	Typically, this kind of methods directly generate markup~(HTML or LaTeX) sequence from raw table image to represent the arrangement of rows/columns and the type of table cells. Originally, long short-term memory (LSTM) networks were used as the decoder under the encoder-decoder framework~\cite{li2020tablebank, end2end1}. TableFormer~\cite{nassar2022tableformer} replaces the LSTM network with transformer and utilizes an additional decoder to realize the end-to-end prediction of. Moreover, EDD~\cite{end2end1} introduces a cell content decoder receiving the information from the structure decoder to boost the recognition performance.

	While the existing methods have achieved promising results on simple or relatively complicated tables, they are still limited in the face of complex TSR problems. Different from above methods mostly depending on one single table component, our proposed GrabTab employs both the features of table basic elements and row-column relations to enhance the representation of cell separators, thus adaptively leveraging multiple components to accommodate such complex scenarios.
	\section{Methodology}
	\subsection{Overall Architecture}
	Intuitively, table separator is the most evident and straightforward visual clue, which is also the basic ingredients of the final output results. Based on this intuition, our GrabTab thus treats it as chief component to dynamically ``grab'' informative clues from other candidate components (the table elements~(orange boxes) and their row/column relations~(connected by blue and purple solid lines)) during deliberation. The architecture of our proposed GrabTab is designed as Fig.~\ref{fig:arch}, which consist of four stages. Given a complex table image, firstly, the candidate components are extracted as element tokens (in orange color) and relation tokens (in green color). Then, the feature of table image along with relation bias is sent to the newly proposed Separator Perceiver~(SP) to obtain separator tokens, which can generate a set of explicit~(red solid lines in Fig.~\ref{fig:arch}) and implicit separator (blue dashed lines) proposals through least squares fitting~(LSF). Afterwards, Components Correlator~(CC) correlates to the separator tokens~(in blue color) with relation and element tokens to obtain the enhanced separator tokens (in purple color). In the end, Structure Composer~(SC) selects the desired separators by predicting their indexes in a sequential manner and re-assembles them as closure cells. The framework is end-to-end trainable by the proposed ``separator losses'' and ``structure loss'', which ensures the versatility of our GrabTab. 
	
	\subsection{Candidate Components Extraction}\label{sec:CCE}
	As aforementioned, our GrabTab extracts table elements and their relations as candidate components, which is expected to provide useful information for the chief separator component. To achieve this goal, we inherit the relation extraction method from a off-the-shell work, NCGM~\cite{liu2022neural}. Specifically, for $N$ table elements, the ``collaborative graph embeddings'' output by NCGM is employed as element tokens in our GrabTab: $\mathbf{E}_{ele} = \left\lbrace \mathbf{{e}}_1, \mathbf{{e}}_2, ..., \mathbf{{e}}_N \right\rbrace \in \mathbb{R}^{N \times d_e} $. Correspondingly, the relation tokens $\mathbf{E}_{rel}$ are also obtained according to the binary-class relations $\left\lbrace \mathbf{R}_{row}, \mathbf{R}_{col.} \right\rbrace$ predicted by NCGM. In details, $\mathbf{R}_{row} = \left\lbrace \mathbf{{r}}_{1,1}, \mathbf{{r}}_{1,2}, ...,\mathbf{{r}}_{i,j},..., \mathbf{{r}}_{N,N} \right\rbrace \in \mathbb{R}^{N^2 \times 2}$, where $\mathbf{{r}}_{i,j} = 1$  if the pair of \textit{i}-th and \textit{j}-th element belong to the same row, and it equals to 0 otherwise. $\mathbf{R}_{col.}$ is denoted in the same manner. To avoid the costly computational consumption brought by element pairs in large amount, according to $\left\lbrace \mathbf{R}_{row}, \mathbf{R}_{col.} \right\rbrace$, we link elements with same relationship as one instance class, \ie, $\mathbf{E}_{row} = \left\lbrace \mathbf{f}_1,  \mathbf{f}_2, ... ,\mathbf{f}_M \right\rbrace \in \mathbb{R}^{M \times d_e} , 
	\mathbf{E}_{col} = \left\lbrace \mathbf{g}_1,  \mathbf{g}_2, ... ,\mathbf{g}_P \right\rbrace \in \mathbb{R}^{P \times d_e}$. Mathematically, for $i$-th row relation instance: $\mathbf{f}_i = \sum_{m=1}^{n}  \mathbf{{e}}_m +  \mathbf{w}_i$, where $\mathbf{w}_i \in \mathbb{R}^{d_e}$ is the $i$-th instance index embedding produced by method~\cite{mikolov2013efficient}. The dictionary size is set to 200 in default. And the $\mathbf{E}_{col}$ is obtained in the similar way. Finally, the relation tokens $\mathbf{E}_{rel} \in \mathbb{R}^{((M+P) \times d_e)}$ are generated as $\mathbf{E}_{rel} = \left\lbrace  \mathbf{E}_{row},  \mathbf{E}_{col}\right\rbrace $.

	\subsection{Separator Proposals Generation}\label{sec:spg}
	\begin{figure}[ht]
		\begin{center}
			\includegraphics[width=1\linewidth, height=0.6\linewidth]{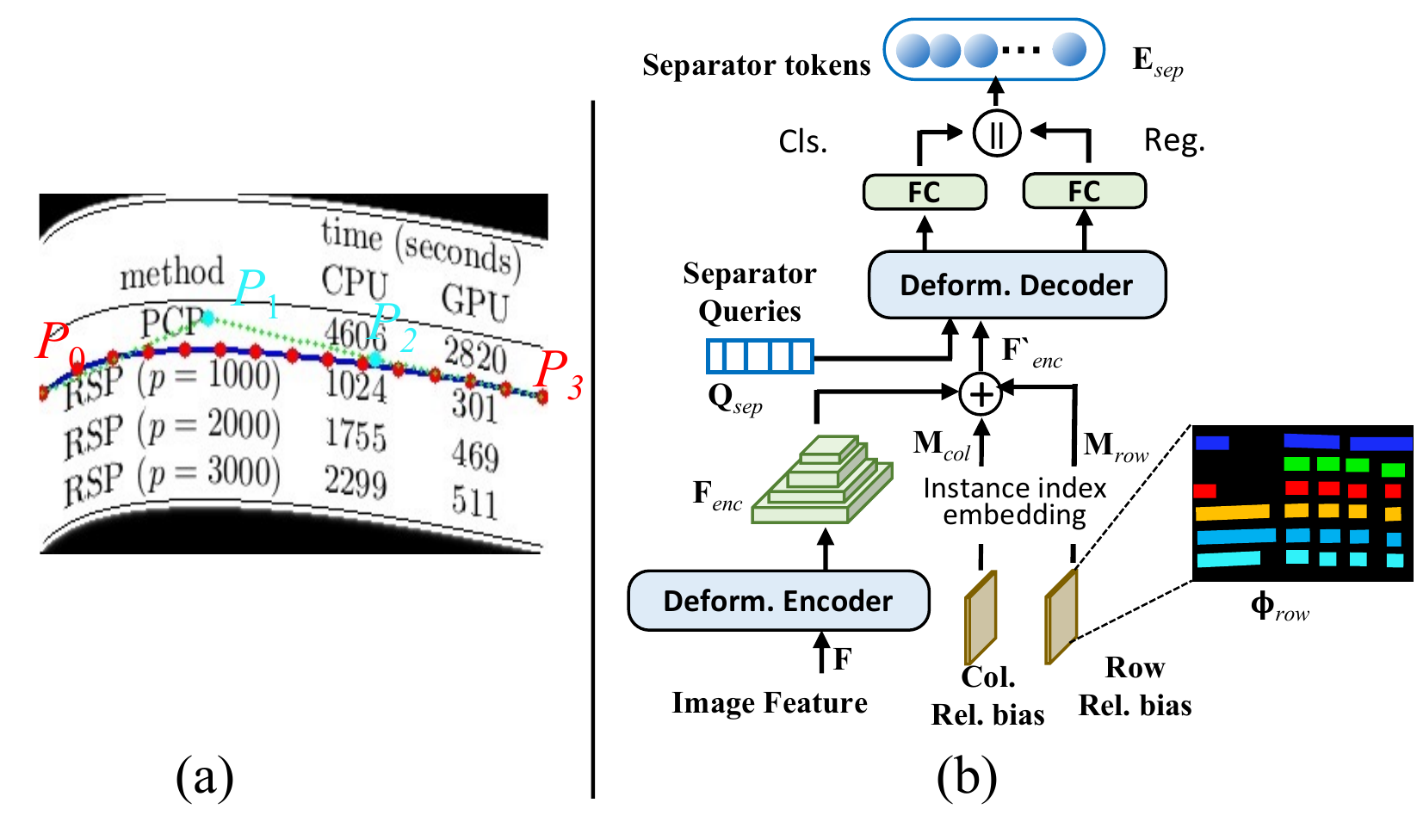}
			\\
			\vspace{-4mm}
		\end{center}
		\caption{(a)~Separator representation. (b)~Separator Perceiver. Best viewed in color.
		}\label{fig:SP}
		\vspace{-3mm}
	\end{figure}
	\noindent\textbf{Separator representation}.\quad Our aim of this stage is to obtain a set of separator proposals, including both explicit and implicit ones shown in Fig.~\ref{fig:arch}, which can well cover the potential cell boundary regions. However, most of previous methods represent cell boundaries with straight lines, which could not fit them properly under distortion scene~(see Fig.~\ref{fig:moti}(b) and (c)). Inspired by work~\cite{liu2020abcnet}, we directly deploy the cubic Bézier curve to represent the curved separators, which is defined as:
	
	\begin{align*}
		\small
		\mathcal{B}(t) &= P_{0}(1-t)^{3} + 3P_{1}t(1-t)^{2}\\
		&+ 3P_{2}t^{2}(1-t) +P_{3} t^{3}, t\in[0,1],
	\end{align*} 
	where $P_{0}$,  $P_{1}$, $P_{2}$  and $P_{3}$ denote control points of the Bézier curve as shown in Fig.~\ref{fig:SP}(a). Intuitively, a straightforward way to produce curved separator is to directly predict the curvature control points. However, it may suffer from instable training procedure and ``point drift'' problem, \ie, small point prediction error can lead to the whole curve misalignment. Alternatively, considering the Bézier curve is parameterized in terms of $t\in[0,1]$, we sample the curve uniformly from $t$ spaced set to get $T$ sample points $\{\mathcal{S}_i({x_i}, {y_i})\}_{i=1}^T$~(red points in Fig.~\ref{fig:SP}(a)), where curve length between each of adjacent are equal. Hence, our method tries to regress these sample points. Once the points determined, the control points of whole curve can be easily acquired by standard ``least squares fitting'' algorithm~(see supplementary materials).   
	
	\noindent\textbf{Separator Perceiver.}\quad Now, we elaborate on how to generate separator proposals with our newly designed Separator Perceiver~(SP). Specifically, as illustrated in Fig.~\ref{fig:SP}(b), considering the scale variety of separators, we build SP upon Deformable DETR~\cite{zhu2020deformable}, where Deformable encoder takes table image feature $\mathbf{F}$ as input. Here, we adopt the canonical ResNet-50~\cite{he2016deep} as the image feature extractor. Then, the multi-scale encoder feature $\mathbf{F}_{enc} = \{D_1, D_2, D_3, D_4\} $  is output, which respectively have strides of {8, 16, 32, 64} pixels with respect to the raw table image in $H$ height and $W$ width. To facilitate the learning of separator, the relation component introduced in Sec.~\ref{sec:CCE} is converted to the column and row relation biases ($ \mathbf{M}_{\sim}\in \left\lbrace \mathbf{M}_{row}, \mathbf{M}_{col}\right\rbrace$) applied on the encoder feature $\mathbf{F}_{enc}$. Taking the row relation bias for example, the row relation mask $\Phi_{row} \in \mathbb{R}^{W \times H}$ is firstly generated according to $\mathbf{R}_{row}$. To be specific, if two elements belong to the same row relations, the pixels inside the element bounding boxes will be assigned with the same relation instance integral index $r, r\in{1,2,...,M}$ (denoted by the same color in Fig.~\ref{fig:SP}(b)), while regions outside element boxes are assigned 0 value.  After down-sampled to the size of each scale feature, we embed these multi-scale maps by embedding method~\cite{mikolov2013efficient} to obtain $\mathbf{M}_{row}$. Then, both $\mathbf{M}_{row}$ and  $\mathbf{M}_{col}$ are added to each scale of encoder feature as $\mathbf{F}'_{enc}$~(see supplementary materials for more details), which is sent to the deformable decoder module subsequently. For the decoder, we adopt $Q$ learnable separator queries $\mathbf{Q}_{sep} \in \mathbb{R}^{Q \times d_q}$ to predict a set of sample points $\{\mathcal{S}_i({x_i}, {y_i})\}_{i=1}^T$ representing each separator. Here, the feature output by corresponding FC~(Fully-Connected) layers of sample points regression and classification are concatenated as separator tokens $\mathbf{E}_{sep} \in \mathbb{R}^{Q \times (2\cdot T + 3)}$, where $2\cdot T$ channels correspond to the x/y coordinates of T sample points while $3$ channels correspond to classes of explicit separators, implicit separators and background.

	\subsection{Multiple Components Correlation}
	\begin{figure}[ht]
		\begin{center}
			\includegraphics[width=1\linewidth,height=0.65\linewidth]{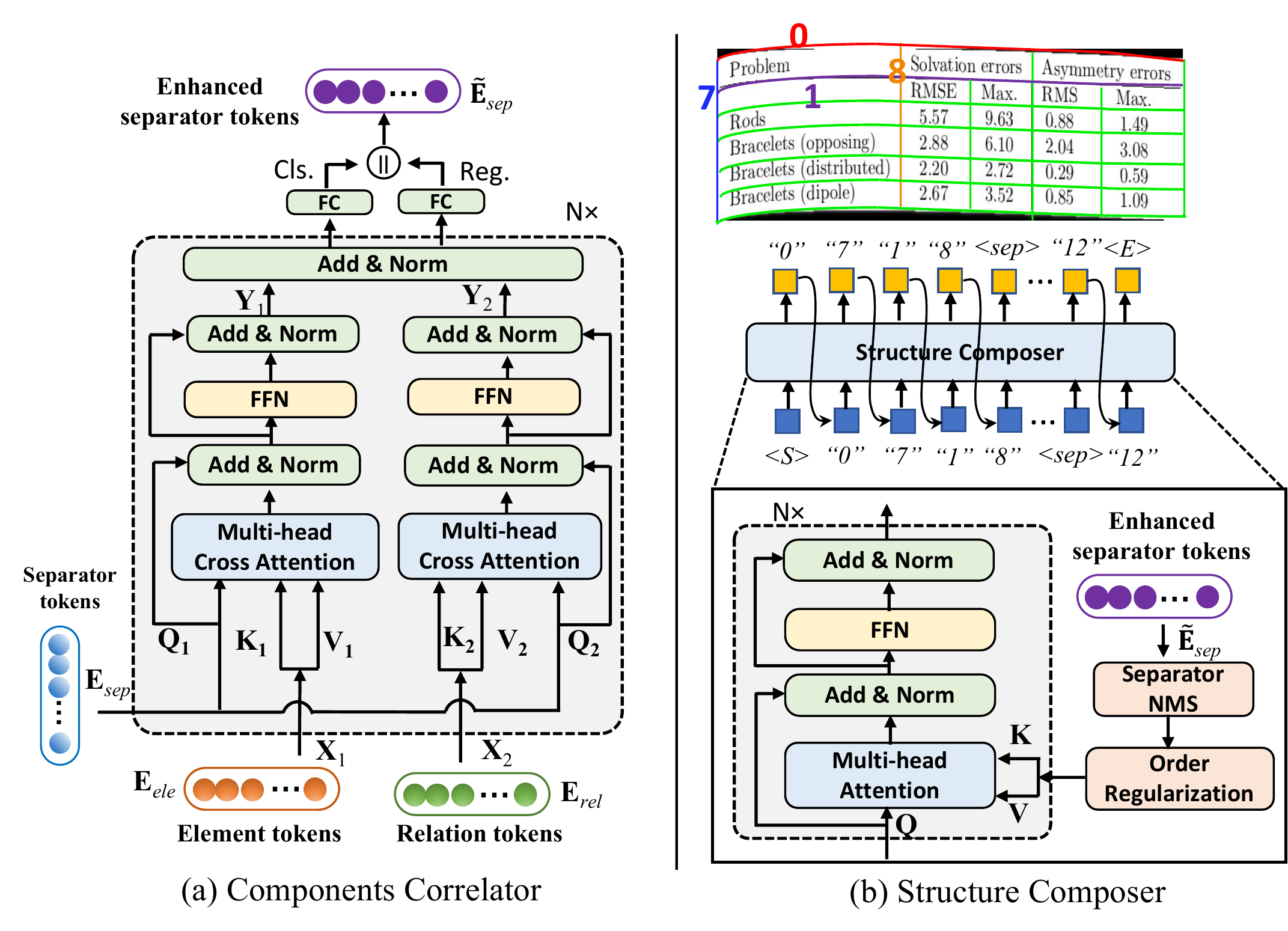}\\
			\vspace{-4mm}
		\end{center}
		\caption{The proposed Components Correlator and Structure Composer. Best viewed in color.
		}\label{fig:CC_SC}
		\vspace{-1mm}
	\end{figure}
	During this phase, the aim is to correlate the candidate components $\mathbf{E}_{ele}$ and  $\mathbf{E}_{rel}$ to the main clue $\mathbf{E}_{sep}$. As demonstrated in Fig.~\ref{fig:CC_SC}(a), to implement above purpose, we build our Components Correlator~(CC) upon the canonical transformer ~\cite{vaswani2017attention} consisting of FeedForward Network~(FFN) and Multi-head Cross Attention~(MHCA) interleaved with Layer Normalization and residual connection. Different from the vanilla one, our CC aggregates the both candidate components in a decoupled way to avoid the interference between the information of large difference they carry. More concretely, the $\mathbf{E}_{sep}$ play as roles of queries~($\mathbf{Q}_{1}$ and $\mathbf{Q}_{2}$) sent to the both streams, where each of them treats either $\mathbf{E}_{ele}$ or  $\mathbf{E}_{rel}$ as the key and value for the MHCA. The block is stacked by $N$ times. Finally, similar to the separator prediction in Sec.~\ref{sec:spg}, we also append the 3-dimension classification and $2T$-dimension regression layers, where the respective FC features are concatenated as the enhanced separator tokens~$\mathbf{\widetilde{E}}_{sep}$. Note, all the queries and keys and values are added with position encoding~(PE)~\cite{vaswani2017attention}, which are omitted in Fig.~\ref{fig:CC_SC}(a) for brevity. Through this way, the separator proposals represented by separator tokens could be refined and aggregated with appropriate exclusive components, which is essential to the final structure prediction.  
	
	\subsection{Table Structure Composing}
	At the final stage of ``deliberation'' process, the refined separator proposals are treated as the ingredients to compose the final table cell in logical order. Motivated by Pix2Seq~\cite{chen2021pix2seq}, we propose a novel generative Structure Composer~(SC) to define above process as a sequence-to-sequence generation problem, based on a intuition that if a model knows about where and what the separators are, we just need to teach it how to compose them as cells. As shown in Fig.~\ref{fig:CC_SC}(b), instead of directly predicting the cell coordinates widely-adopted in most methods, our SC predicts a sequence of selected separator indexes where each four corresponding separators construct a cell's boundary. 
	
	Formally, the output sequence $\mathcal{G}$ is represented as $\{ \langle S \rangle, C_0,\langle sep \rangle, C_1, ..., C_g, \langle E \rangle\}$, where $C_{\sim} = [\alpha_{top}, \alpha_{left}, \alpha_{bottom}, \alpha_{right}]$ are the anticlockwise arranged indexes of separators wrapping a cell. $\langle S \rangle$, $\langle sep \rangle$ and $\langle E \rangle$ are start, separation and end tokens respectively. Given a start token ``$\langle S \rangle$'' as query, the SC recursively predicts index sequence $\mathcal{G}$ until the ``$\langle E \rangle$'' is output. The basic block of SC is also built on the vanilla transformer block~\cite{vaswani2017attention}, where enhanced separator tokens $\mathbf{\widetilde{E}}_{sep}$ are regarded as keys and values with PE added. 
	
	To control the input sequence length to a reasonable account, before sent to the transformer blocks, $\mathbf{\widetilde{E}}_{sep}$ and corresponding separator proposals are firstly processed by the ``Separator NMS''. As the original NMS~\cite{ren2015faster} is designed for the object detection task, to adapt it to the TSR task, we repurpose it into our ``Separator NMS'' by replacing the IoU metric with our proposed Separator Distance: 
	\begin{align}
		\small
		\mathcal{D}_{sep}  = &\frac{1}{T} \sum_{t \in T} ||\mathcal{S}^{(i)}_t(x_t,y_t) - \mathcal{S}^{(j)}_t(x_t,y_t)||_{2}, \label{eq:sd}
	\end{align}	
	where $\mathcal{S}^{(i)}_t(x_t,y_t)$ is the $t$-th sample points from the $i$-th separator~(corresponds to the $i$-th token of $\mathbf{\widetilde{E}}_{sep}$). The condition for removing separator is set as $\mathcal{D}_{sep} < \sigma$, where $\sigma$ is set to 5 in default. Considering the selected separators are unordered, which could cause training collapse problem, we sent the selected separators to the ``Order Regularization'' submodules. They are firstly grouped into horizontal~($\mathcal{H}_h$) and vertical~($\mathcal{H}_v$) types according to the slope of the first~(start) sample point and last~(end) one connected line. The output of ``Order Regularization'' is defined as:
	\begin{align}
		\small
		\mathcal{H} = \{Sort_y(\mathcal{H}_h), Sort_x(\mathcal{H}_v)\},
	\end{align}     
	where ``$Sort_x$'' and ``$Sort_y$'' denotes sort operations along x and y axes respectively.
	
	\subsection{Training Strategy}\label{sec:TS}
	
	\noindent\textbf{Loss function.}\quad As illustrated in Fig.~\ref{fig:arch}, our proposed GrabTab is trained in an end-to-end way by multi-task
	loss $\mathcal{L} = \lambda_{1}\mathcal{L}_{sep_1} + \lambda_{2}\mathcal{L}_{sep_2} +\lambda_{3}\mathcal{L}_{struct}$, where balance weight of each loss $\lambda_{\sim}$ is set to 1 equally. Among, the separator loss $\mathcal{L}_{sep^{\sim}}$ is the modified Hungarian matching loss~\cite{zhu2020deformable} adapting to the TSR task: 
	\begin{align}
		\small
		\mathcal{L}_{sep_{\sim}}(O, G) = \min_\gamma\underbrace{
			\left( \sum_{i=1}^M \mathcal{L}_{cls}(c_{\gamma_i}, \hat c_i) + \mathcal{L}_{reg}( s_{\gamma_i}, \hat s_i)\right)}
		_{\mathcal{L}(O, G | \gamma)}.\label{eq:seploss} 
	\end{align}   
	Similar to \cite{zhu2020deformable}, during training, the model firstly assigns each output $O=\{ s_j,  c_j\}_{j=1}^{M}$ to exactly one annotated separator in $G=\{\hat s_k, \hat c_k\}_{k=1}^{2}$ or background $\emptyset$, where $c_k$ is the $k$-th class~(explicit or implicit separator). Note that $O$ and $G$ are both denoted by sample points. $\gamma_i \in \{\emptyset, 1, 2\}$ is the assignment of model output $i$ to ground truth~(GT) $\gamma_i$, while $\mathcal{L}_{reg}$ is implemented by Eqn.~\eqref{eq:sd}. For the 
	$\mathcal{L}_{struct}$, we adopt the standard cross-entropy loss which is similar with other generative models~\cite{chen2021pix2seq}~(see supplementary materials). 
	
	\noindent\textbf{Separator assignment.}\quad Nevertheless, the above one-to-one assignment in vanilla Hungarian matching algorithm would cause the severe separator missing problem, due to the slim shape of separator. To attack it, inspired by method~\cite{ouyang2022nms}, we further modify the assignment to the one-to-many strategy, \ie, one GT is assigned with a group rather one separator. Based on the one-to-one assigned separator $\left\lbrace s_{\gamma_i}, c_{\gamma_i} \right\rbrace$, we further find its neighboring ones on the condition that $\mathcal{D}_{sep}<5$. Here, $\mathcal{D}_{sep}$ is the Separator Distance (defined in Eqn.~\eqref{eq:sd}) between $\left\lbrace s_{\gamma_i}, c_{\gamma_i} \right\rbrace$ and its neighboring separators. Afterwards, the grouped separators are also assigned to the GT annotation $\gamma_i$.
	

	\section{Experiments}
	\subsection{Datasets and Evaluation Protocol}
	\noindent\textbf{Datasets.}\quad
	We evaluate our method on the following benchmark datasets under both complex and regularized table scenarios. ICDAR-2013~\cite{gobel2013icdar}, ICDAR-2019~\cite{gao2019icdar}, WTW~\cite{long2021parsing}, UNLV~\cite{shahab2010open}, SciTSR~\cite{chi2019complicated}, SciTSR-COMP~\cite{chi2019complicated} and SciTSR-COMP-A~\cite{liu2022neural} are evaluated  under protocol of physical structure recognition, while TableBank~\cite{li2020tablebank} and PubTabNet~\cite{zhong2020image} are adopted to evaluate the logical structure recognition performance. Furthermore, WTW~\cite{long2021parsing}, SciTSR-COMP~\cite{chi2019complicated} and SciTSR-COMP-A~\cite{liu2022neural} are employed as complex TSR datasets with more challenging distractors involved, while the rest are the regularized ones. Supplementary material presents more details on them. 
	
	\noindent\textbf{Evaluation protocol.}\quad
	For a fair comparison, we inherit the widely-adopted protocols from prevalent methods. Among, precision, recall and F1-score are utilized to evaluate the performance of recognizing table physical structure. And the performance of table logical structure recognition is evaluated by the Tree-Edit-Distance-based Similarity (TEDS)~\cite{zhong2020image} and BLEU score~\cite{papineni2002bleu} protocols.  
	
	\subsection{Implementation Details}
	We build the framework using Pytorch~\cite{paszke2019pytorch} and conduct all experiments on a workstation with 8 Nvidia Tesla V100 GPUs. All the component tokens are projected into 256-dimensional vertors. All the transformer block numbers in SP, CC and SC are set to 6, where the dimensions of hiddens and FFN are 256 and 2,048 respectively. The number of queries in SP is empirically set to 1,000. The framework is optimized by AdamW~\cite{loshchilov2017decoupled} with a batch size of 16. We adopt the learning rate $1\mathrm{e}{-5}$ for both Seperator Perceiver and Components Correlator, and $1\mathrm{e}{-4}$ for Structure Composer. The number of sample points $T$ for representing separator is set to 15. During the training phase, the table images within a same batch are randomly resized ranging from 480 to 800 while the size is fixed to 1,100 for test. For all experiments, the network is pre-trained on SciTSR for 10 epochs, and then fine-tuned on different benchmarks for 50 epochs.
	
	\begin{table*}[htb!]
		\centering
		\small
		\setlength{\tabcolsep}{0.5mm}
		\scalebox{1} {
			\begin{tabular}{l|l|ccc|l|ccc|l|ccc|l|c|l|c}
				\toprule[1.5pt]
				&\multicolumn{4}{c|}{\textbf{WTW}} &\multicolumn{4}{c|}{\textbf{SciTSR-COMP}} &\multicolumn{4}{c|}{\textbf{SciTSR-COMP-A}} &\multicolumn{2}{c|}{\textbf{TableBank}} &\multicolumn{2}{c}{\textbf{PubTabNet}} \\
				\hline
				Method& \tabincell{l}{Train Set} & P & R & F1 & \tabincell{l}{Train Set} & P & R & F1& \tabincell{l}{Train Set} & P & R & F1& \tabincell{l}{Train Set} & BLEU& \tabincell{l}{Train Set} & TEDS\\
				\hline
				{C-CTRNet~\cite{long2021parsing}} & \tabincell{l}{WTW} & 93.3 &91.5 &92.4 &-&-&-&-&-&-&-&-&-&-&-&-\\
				FLAG-Net~\cite{liu2021show} &WTW& 91.6&89.5&90.5 &SciTSR & 98.4 &98.6 &98.5 & Sci. + Sci.-C-A & 82.5 &83.0 &82.7&SciTSR & 93.9&SciTSR & 95.1\\
				TSRFormer~\cite{lin2022tsrformer}  & WTW & 93.7 & 93.2 & 93.4  &SciTSR & 99.1 & 98.7 & 98.9&-&-&-&-&-&-&-&-\\
				NCGM~\cite{liu2022neural} & \tabincell{l}{WTW} & 93.7 & 94.6 & 94.1 &SciTSR & 98.7 &98.9 &98.8  & Sci. + Sci.-C-A& 88.4 &90.7 &89.5 &SciTSR& 94.6 &  SciTSR & 95.4\\
				\hline
				\textbf{GrabTab} & WTW & \textbf{95.3} & \textbf{95.0 }&  \textbf{95.1} & SciTSR  & \textbf{98.9} & \textbf{99.4} & \textbf{99.1}  &  Sci. + Sci.-C-A & \textbf{94.3} &  \textbf{93.8} &   \textbf{94.0} &SciTSR& \textbf{95.0}&  PubTabNet & \textbf{97.2}\\
				
				\bottomrule[1.5pt]
			\end{tabular}
		}
		\caption{Comparison results on WTW, SciTSR-COMP, SciTSR-COMP-A, TableBank and PubTabNet datasets. ``P'', ``R'' and ``F1'' stand for ``Precision'', ``Recall'' and ``F1-score'' respectively. ``C-CTRNet'' and ``Sci.'' are short for ``Cycle-CenterNet'' and SciTSR.}
		\label{Comparison_results}
		\vspace{-0.2cm}
	\end{table*}
	
	\subsection{Comparison with State-of-the-arts}

	\noindent\textbf{Results of complex table structure recognition.}\quad Tab.~\ref{Comparison_results} gives comparison results on several benchmark datasets. Among, the first three columns indicate the performance on recognizing complex tables containing more severe distractors. Compared with existing methods, the F1-score of our GrabTab can beat the second best method, NCGM~\cite{liu2022neural}, by 4.5\%  on SciTSR-COMP-A dataset, while the apparent performance improvement is also witnessed on WTW and SciTSR-COMP datasets. This phenomenon further confirms that simply focusing on single component extraction is not the optimal solution. By equipped with deliberation mechanism, our GrabTab finds the silver linings behind the problem of multiple components leverage, with decent results produced.
	
	\noindent\textbf{Results of regularized table structure recognition.}\quad
	In the last two columns of Tab.~\ref{Comparison_results}, the performance on regularized table cases is also given. As they are evaluated under logical structure recognition protocol, we convert the output physical structure format to the HTML, which strictly follows the operation in NCGM. From the table, one can observe that, on both datasets, our GrabTab can also achieve the consistent improvement than the state-of-the-arts. More results are given in the supplementary materials. 
	
	\noindent\textbf{Computational complexity.}\quad
	Please refer to the supplementary materials for details.
	
	\subsection{Ablation Study}
	\begin{table}[htb]
		\small
		\setlength{\tabcolsep}{0.6mm}
		\centering
		\begin{tabular}{l|c|c|cc|cc|ccc}
			\toprule[1.5pt]
			Method&	RB & CC & OA & MA  & RP& SC & P & R& F1 \\
			\hline
			{Deform.-DETR} &	\xmark & \xmark & \cmark & \xmark & \cmark & \xmark &84.8 &85.1 &84.9   \\
			GrabT.$_\text{w/o CC/MA/SC}$  &	\cmark & \xmark &	\cmark  & \xmark &	\cmark  & \xmark &88.7 &89.4  &  89.0 \\
			GrabT.$_\text{w/o MA/SC}$ &	\cmark & \cmark & \cmark & \xmark &	\cmark  & \xmark & 90.2 & 90.4& 90.3  \\
			GrabT.$_\text{w/o SC}$ &	\cmark & \cmark & \xmark & \cmark &	\cmark  & \xmark  & 91.5 & 91.8 & 91.6  \\
			\hline
			{GrabTab-S} &	\cmark & \cmark & \xmark & \cmark & \xmark & \cmark & {91.9} & {92.4}  & {92.1} \\
			\textbf{GrabTab} &	\cmark & \cmark & \xmark & \cmark & \xmark & \cmark & \textbf{94.3} & \textbf{93.8}  & \textbf{94.0} \\
			\bottomrule[1.5pt]
		\end{tabular}  
		\caption{Ablation studies of GrabTab on SciTSR-COMP-A dataset. Legend: ``RB'': Row/Column Relation Bias, ``CC'': Components Correlator, ``OA'': One-to-One Assignment, ``MA'': One-to-Many Assignment, ``RP'': Rule-basd Post-processing, ``SC'': Structure Composer. ``w/o.'' is short for ``without'' and ``-S'' represents straight-line separators.}
		\label{ablation}
	\end{table}
	
	In this subsection, we investigate the effects of various factors in GrabTab by juxtaposing analytic experiments on SciTSR-COMP-A dataset, which is the most challenging complex dataset. If we remove all the tailored designs in our GrabTab, it would degenerate to the ``Deform.-DETR'' regarding separators as prediction targets. By simply adding relation biases~($ \mathbf{M}_{\sim}\in \left\lbrace \mathbf{M}_{row}, \mathbf{M}_{col}\right\rbrace$ in Sec.~\ref{sec:spg}), we surprisingly observe the 4.1\% F1-score improvement brought by ``GrabT.$_\text{w/o CC/MA/SC}$''. Moreover, if the Components Correlator~(CC) submodule is appended, the performance could be further boosted by ``GrabT.$_\text{w/o MA/SC}$'', which is a persuasive evidence to demonstrate the effectiveness of CC. 
	
	In addition to relation biases~(RB) and CC, separator assignment way is another factor of importance. By adopting original one-to-one assignment~(OA) trick, the F1 performance witnesses 1.3\% drop compared with the version~(``GrabT.$_\text{w/o SC}$'') equipped with our modified one-to-many assignment~(MA) introduced in Sec.~\ref{sec:TS}. We attribute the performance drop to the ``separator missing'' problem, \ie, most separators are not perceived under original one-to-one assignment.
	
	As elaborated above, most existing methods depend on complicated post-processing based on heuristic, which is not versatile for various complex tables. Comparatively, our Structure Composer~(SC) is able to learn how to pick up desired separators and compose them into the final structure according to the specific cases. As indicated by ```GrabTab'', which is the full version of our method, by equipping it with SC, the F1-score can be increased to 94.0\%, which surpasses the rule-based version~(``GrabT.$_\text{w/o SC}$'') by a large margin. To investigate the effect of separator quality, we further modify the prediction targets in ``GrabTab-S'', where target separators are represented in straight-line format instead. Consequently, we find the modification is detrimental to the performance, which can be attributable to the inconsistence between target curved representation and predicted straight lines, especially highlighted for the distorted tables.

	\begin{figure*}[htb]
		\includegraphics[width=0.98\linewidth,height=0.37\linewidth]{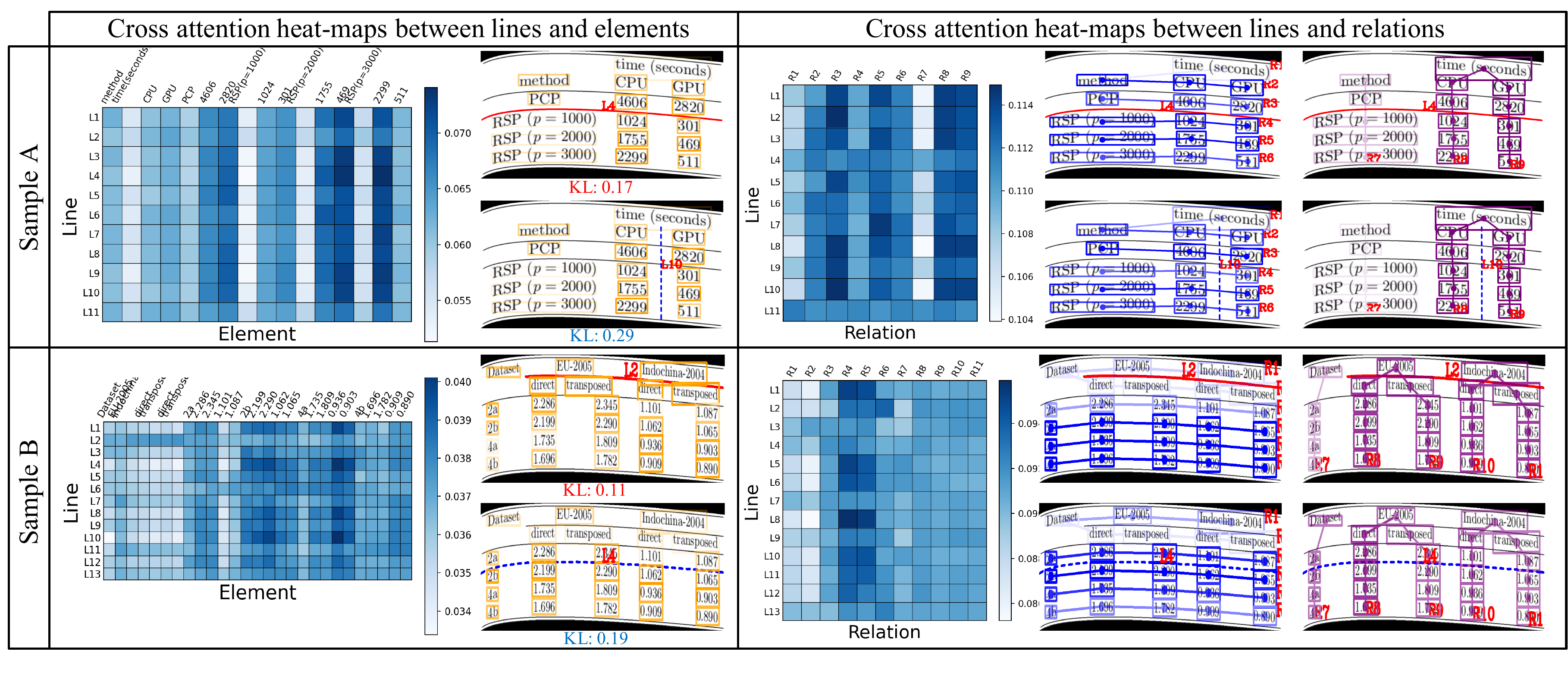}
		\vspace{-0.2cm}
		\caption{Visualizations of the attention heat-maps from last block of Components Correlator. Table elements or relation connections with darker colors indicate stronger correlations with exemplar lines. Best viewed in color and zoomed in. 
		} 
		\label{fig:heatmap}
	\end{figure*}

	\subsection{Further Analysis on Deliberation}
	\noindent\textbf{What components matter during deliberation?}\quad
	To investigate the behaviors of different components during deliberation, for two samples from SciTSR-COMP-A dataset, we in Fig.~\ref{fig:heatmap} visualize the attention heat-maps from last block of Components Correlator module to reflect the correlations between separator lines and candidate components, \ie, elements~(orange boxes) and their relations of row (blue connections)/column~(purple connections). For clarity, for each sample, we pick up one explicit~(red solid curve) and implicit~(blue dashed curve) separator line as examples, and directly draw the correlations on the raw table images. The darker color indicates stronger correlation.

	As demonstrated in the left part of Fig.~\ref{fig:heatmap}, we observe that explicit lines~(``L4'' in ``Sample A'' and ``L2'' in ``Sample B'') can attend to nearly all elements in global range while present smooth distribution of attention weights. In contrast, implicit lines~(``L10'' in ``Sample A'' and ``L4'' in ``Sample B'') are often inclined to establish the relationships with nearby spanning elements, which indicates that adjacent elements can contribute more to recovering invisible implicit lines. To further quantitatively compare the attention patterns between explicit and implicit lines, we employ Kullback-Leibler divergence~\cite{joyce2011kullback} as the measurement. Comparatively, the implicit lines present higher averaged divergence than that of explicit ones, which also confirms that implicit lines have stronger correlation with elements. Thanks to it, the performance can be boosted consistently. 
	
	On the other hand, in order to show the cooperation patterns between lines and relations, the cross attentions between lines/row-relations and lines/column-relations are separately visualized in the right part of Fig.~\ref{fig:heatmap}. Obviously, the explicit lines prefer those relations carrying the same span information ~(``L4'' related ``R2''$\sim$``R9'' in ``Sample A'' and ``L2'' related ``R4'' $\sim$ ``R11'' in ``Sample B''), while the implicit lines pay more attention to the adjacent local relations~(``L10'' related from ``R8'' to ``R9'' in ``Sample A'' and ``L4'' related from ``R3'' to ``R5'' in ``Sample B''), which vividly illustrates the importance of relational features to the reconstruction of invisible separators.  Simultaneously,  the effectiveness of the CC module for ``grabbing'' informative clues between lines and relations is also verified.
	
\begin{figure}[htb]
		\centering
		\subfigure[Sample results of GrabTab on SciTSR-COMP-A dataset.] {
			\includegraphics[width=0.97\linewidth, height=4.4cm]{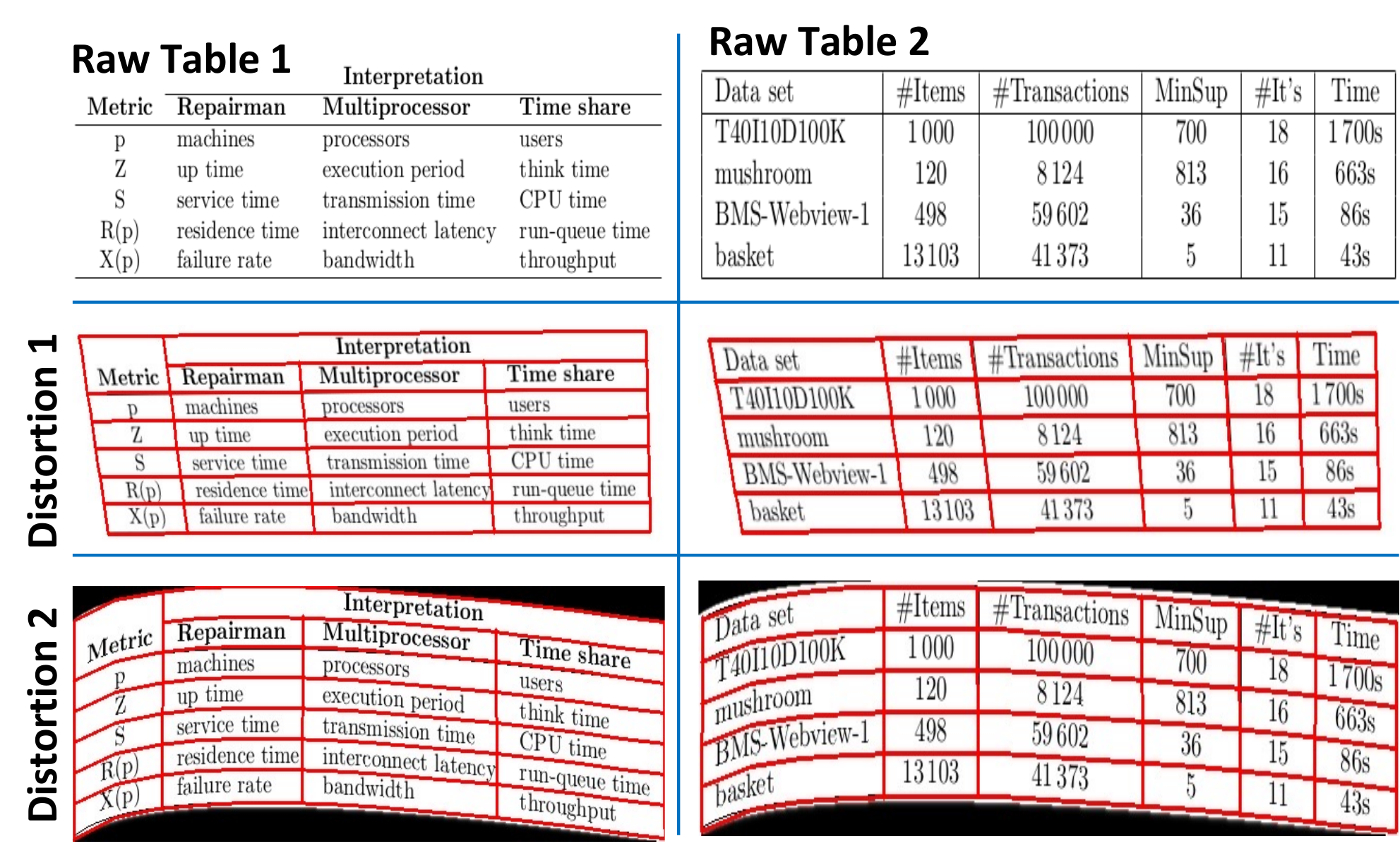}
		}
		\subfigure[Sample results of GrabTab on WTW dataset.] {
			\begin{minipage}[c]{1\linewidth}
				\includegraphics[width=0.47\linewidth, height=2.3cm]{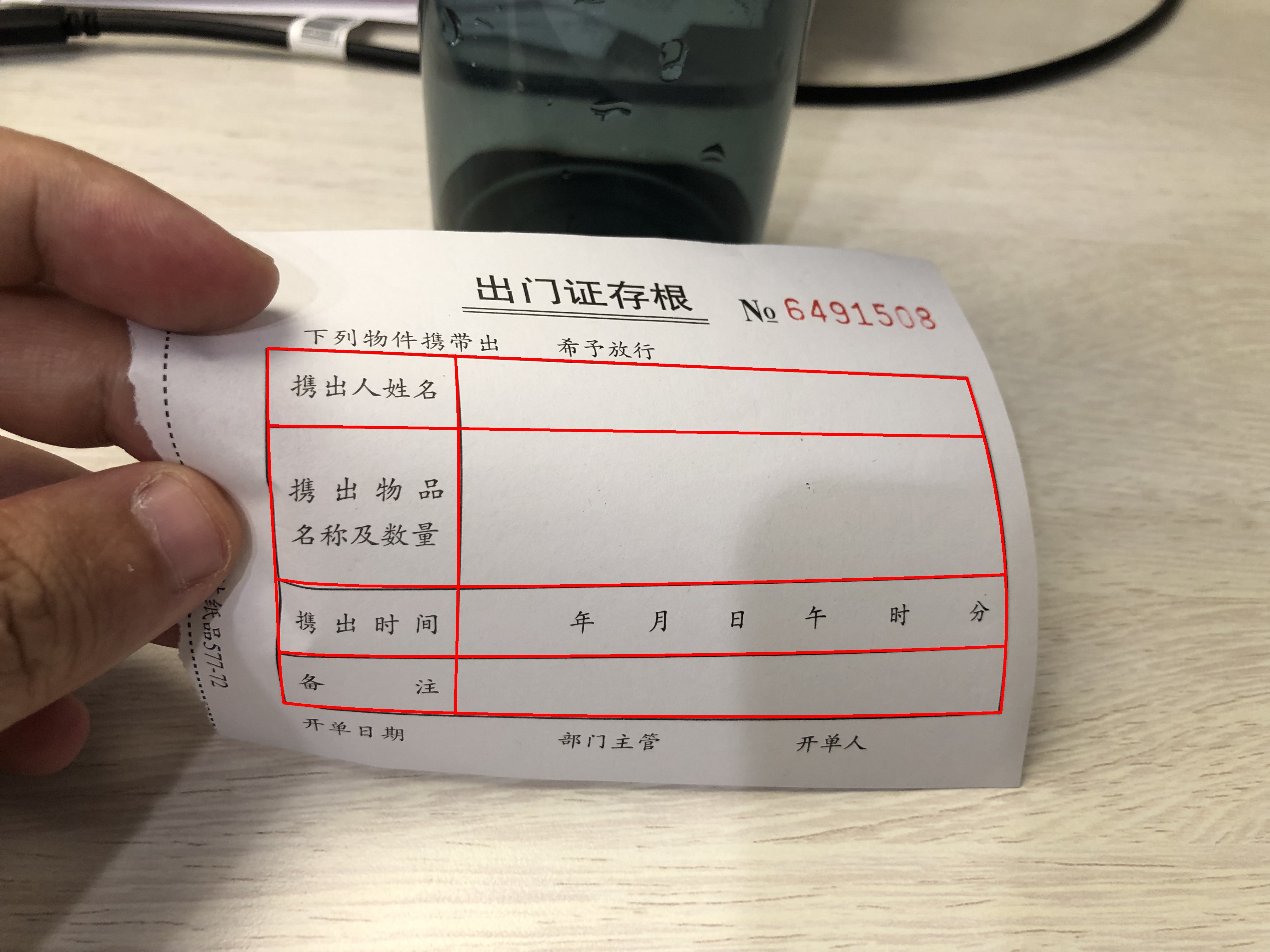}
				\includegraphics[width=0.47\linewidth, height=2.3cm]{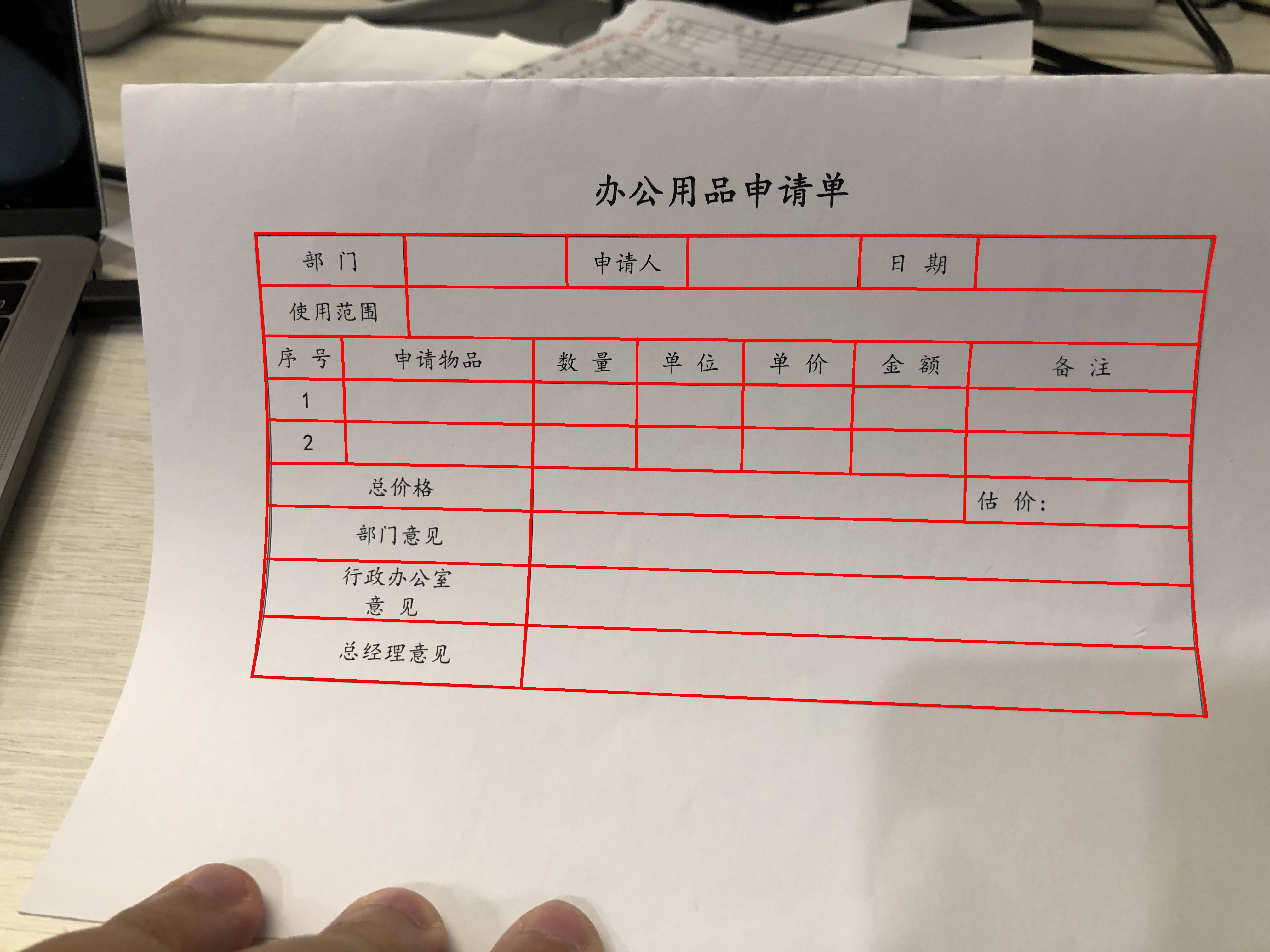}
				\\
				\includegraphics[width=0.47\linewidth, height=3cm]{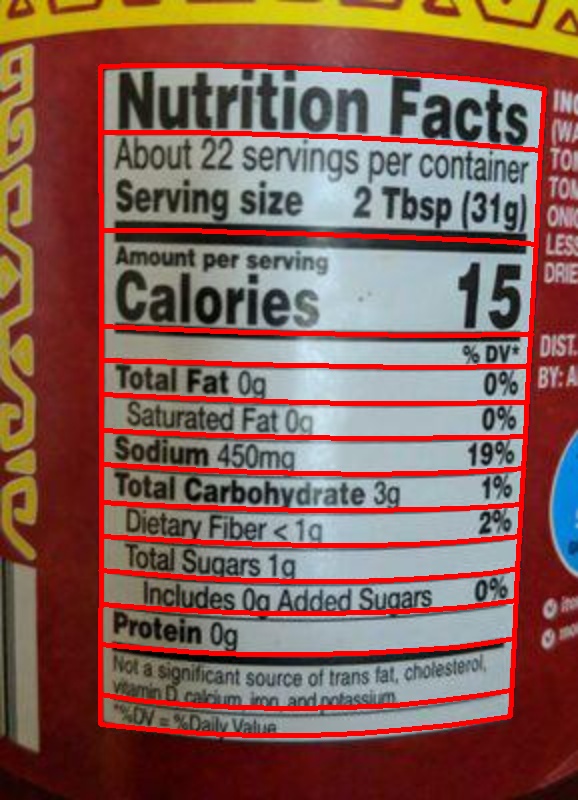}
				\includegraphics[width=0.47\linewidth, height=3cm]{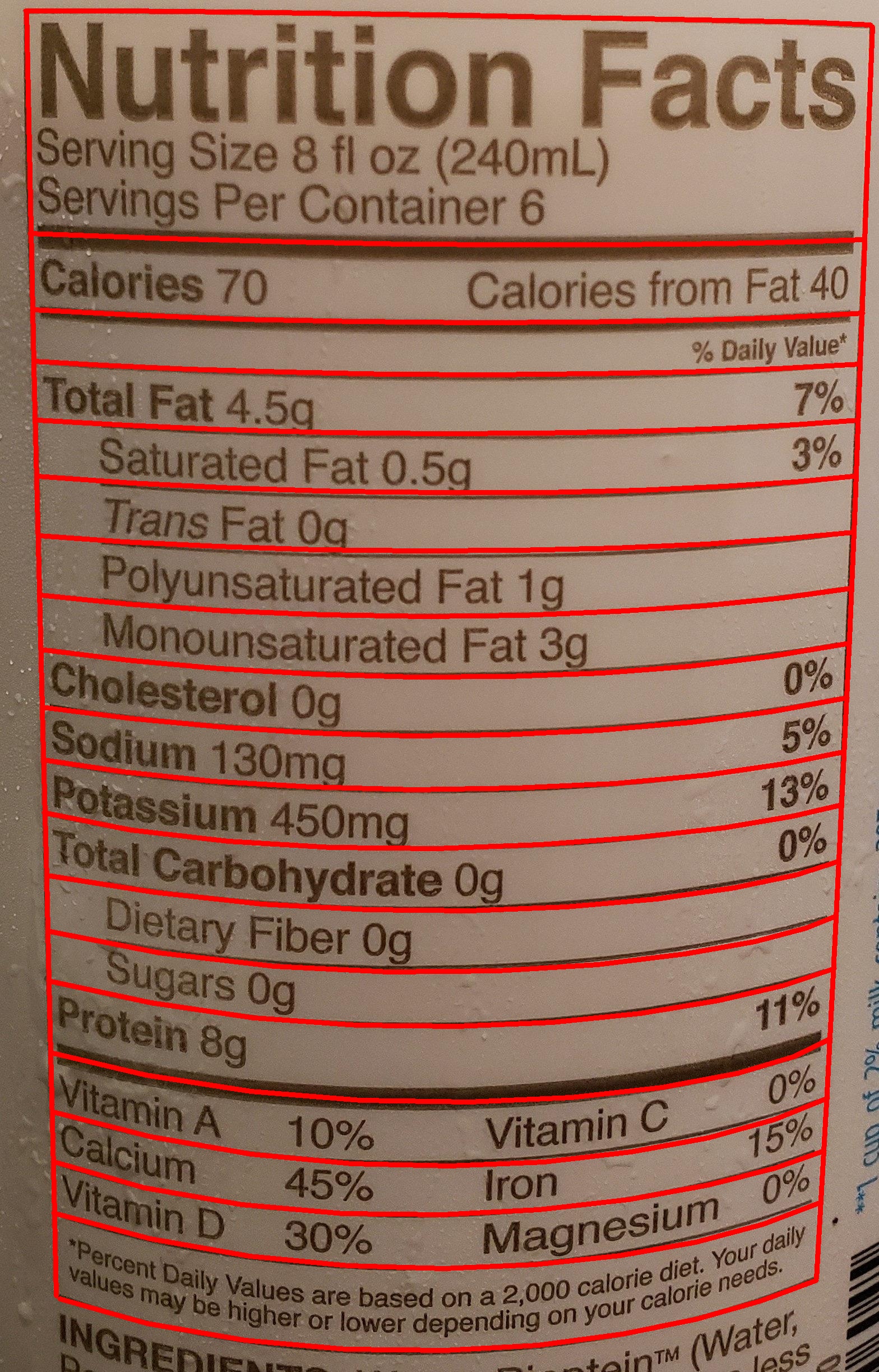}
			\end{minipage}
		}
		\vspace{-0.3cm}
		\caption{Qualitative results on SciTSR-COMP-A and WTW datasets.}
		\label{fig:result} 
		\vspace{-0.5cm}
	\end{figure}

	\noindent\textbf{Qualitative results.}\quad
	We in Fig.~\ref{fig:result} visualize several recognition results from two complex TSR datasets, \ie,  SciTSR-COMP-A~\cite{liu2022neural} and WTW~\cite{long2021parsing}, which further confirm the versatility of our GrabTab. Among, the tables in SciTSR-COMP-A are distracted by two kinds of synthesized distortions. From the results in Fig.~\ref{fig:result}(a), one can observe that our method can precisely predict the distorted table cells, even with severe content misalignment. Moreover, when facing with camera captured tables from real world, where tables often distracted by light change and unpredictable background objects, our method  still exhibits robustness to them and correctly yields the promising results. 
	
	\section{Conclusion}
	In this work, we propose a GrabTab method equipped with deliberation mechanism, which can flexibly deal with complex tables with multiple components assembled but without complicated post-processing involved. This exclusive mechanism ensure the robustness and versatility of our method. We conduct extensive experiments on both complex and regularized table datasets to validate our method and investigate its working pattern. Experimental results demonstrate that our model outperforms prior arts, especially under complex table scene, which verifies the effectiveness of our method.

	\begin{appendices}
		\renewcommand\thesection{\Alph{section}} 
		\renewcommand\thesubsection{\Alph{section}.\arabic{subsection}} 
		\renewcommand\thefigure{\Alph{section}\arabic{figure}} 
		\renewcommand\thetable{\Alph{section}\arabic{table}} 
		\setcounter{section}{0}
		\setcounter{figure}{0}	
		\setcounter{table}{0}
		\section{Preliminaries}
		\subsection{Least Squares Fitting}
		As introduced in the main text, we adopt standard Least Squares Fitting~(LSF) algorithm to obtain each curved separator from a set of predicted sample points $\{\mathcal{S}_i\}_{i=1}^{T}$. As curved separator is represented by the cubic Bézier curve~\cite{liu2020abcnet}, our goal is to obtain control points $\{P_i\}_{i=1}^{3}$ through the LSF defined as:
		\begin{equation}
			\renewcommand\arraycolsep{1.5pt}
			\begin{bmatrix}
				P_{0}\\
				P_{1}\\
				\vdots \\
				P_{3}\\
			\end{bmatrix}
			=
			\begin{bmatrix}
				\mathcal{S}_{x_0} & \mathcal{S}_{y_0}\\
				\mathcal{S}_{x_1} & \mathcal{S}_{y_1}\\
				\vdots & \vdots\\
				\mathcal{S}_{x_T} & \mathcal{S}_{y_T}\\
			\end{bmatrix}
			\begin{bmatrix}
				b_{0,3}(t_0) & \cdots\ & b_{3,3}(t_0)\\
				b_{0,3}(t_1) & \cdots\ & b_{3,3}(t_1)\\
				\vdots&  \ddots\ &  \vdots\\
				b_{0,3}(t_T) & \cdots\ & b_{3,3}(t_T)
			\end{bmatrix}^ \mathrm{ T },
			\label{eq:matrix}
		\end{equation}
		where $\{t_i\}_{i=0}^m$ is uniformly sampled from $0$ to $1$ and $b_{i,n}$ are Bernstein basis polynomials of degree $n$:
		\begin{align}
			\label{eq:bernstein}
			b_{i,n} = C_{n}^{i} t^{i}(1 - t)^{n - i},~i=0,...,n.
		\end{align}
		Here, $n=3$ for the cubic Bézier curve we adopted~(see Sec.~\ref{sec:spg} of main text) . 
		
		\subsection{Multi-head Attention}
		We build the core Components Correlator and Structure
		Composer of our GrabTab upon Multi-head Attention (MHA)~\cite{vaswani2017attention} module. Here, we give a brief introduction on it. Mathematically, given queries $\mathbf{Q}$, keys $\mathbf{K}$ and values $\mathbf{V}$, the definition is given as:
		\begin{align*}
			\small
			\centering
			&MultiHead(\mathbf{Q,K,V}) = Concat(\mathbf{H}_1,\mathbf{H}_2,...,\mathbf{H}_h)\mathbf{W^*},\\
			& \mathbf{H}_i = Attention(\mathbf{QW}_i^Q,\mathbf{KW}_i^K,\mathbf{VW}_i^V ),  i \in \{1,2,..., h\},\\
			&Attention(\mathbf{Q,K,V}) = softmax(\frac{\mathbf{Q}\mathbf{K}^\top}{\sqrt{d_k}})\mathbf{V},\label{eq:att}
		\end{align*}
		where $d_k$ is the keys dimension, and $h$ is the multiple heads number. $\mathbf{W}_i^Q \in \mathbb{R}^{d_m \times d_k},\mathbf{W}_i^K \in \mathbb{R}^{d_m \times d_k},\mathbf{W}_i^V \in \mathbb{R}^{d_m \times d_v}$ and $\mathbf{W}_i^* \in \mathbb{R}^{hd_v \times d_m}$ are matrices of transformation weights individually.
		
		\subsection{Kullback-leibler divergence}
		As elaborated in ``Further Analysis on Deliberation'' subsection of main text, we introduce Kullback-leibler divergence~\cite{joyce2011kullback} (KL-divergence) to measure the averaged  diversity of attention maps from CC. For each separator line, its KL-divergence can be defined as:
		\begin{align}
			&\mathcal{M}_i = \frac{1}{n}  \sum_{j=1}^n 	D_{\mathrm{KL}}(\mathbf{P}_i\| \mathbf{P}_j),\\ 	 
			&D_{\mathrm{KL}}(P \| Q) =\sum_{x \in \mathcal{X}} P(x) \log \left(\frac{P(x)}{Q(x)}\right)
		\end{align}
		where $\mathbf{P}_i$ and $\mathbf{P}_j$ correspond to the $i$-th and $j$-th row attention distribution vector in attention map~(see Fig.~\ref{fig:heatmap} in main text). Note, the attention weight in attention map is obtained by averaging the weights from multiple heads.

		\section{Relation Bias Extraction}
		\begin{figure}[t]
			\begin{center}
				\includegraphics[width=1\linewidth]{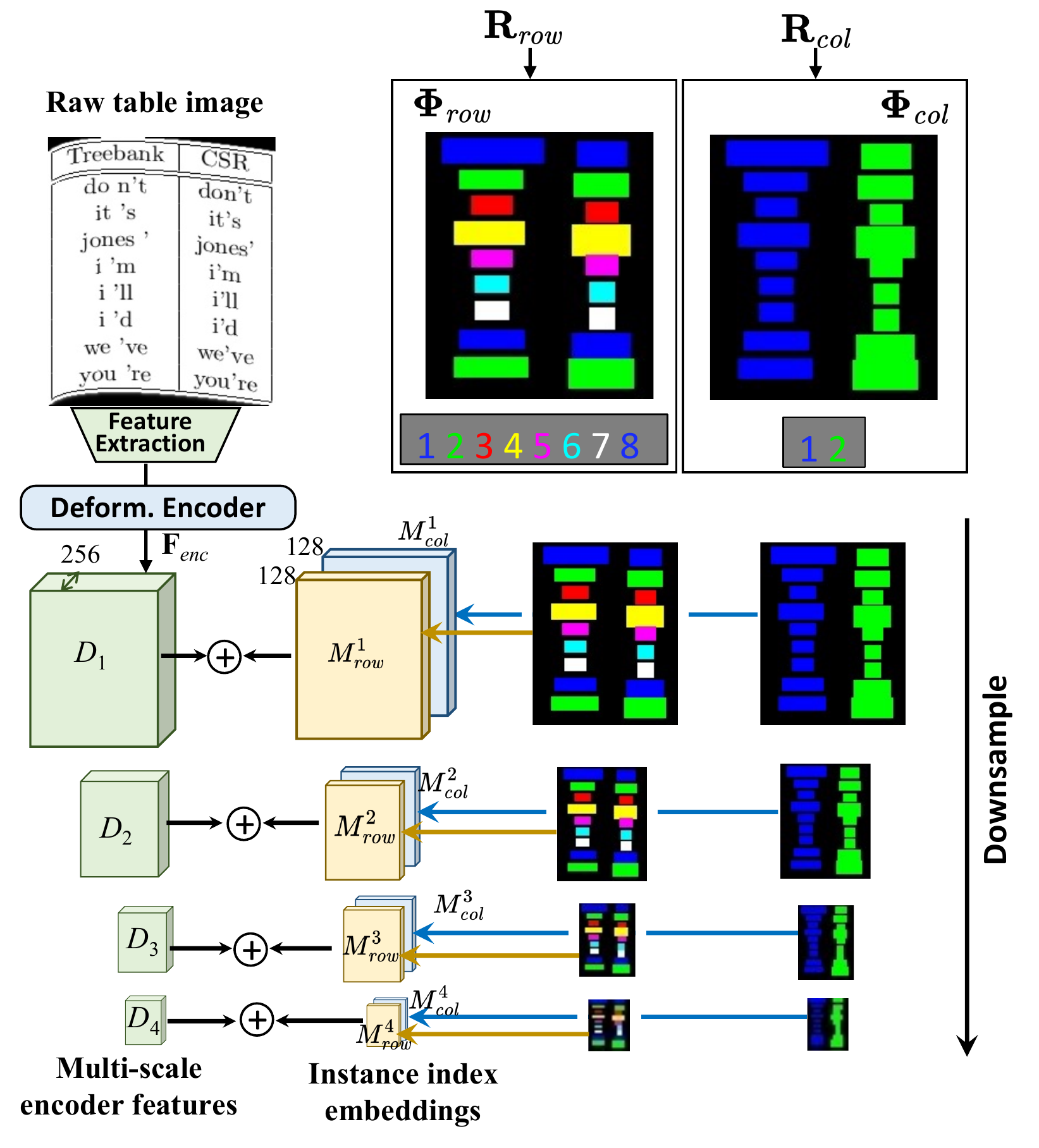}
				\\
				\vspace{-3mm}
			\end{center}
			\caption{Process of relation bias extraction. Best viewed in color and zoomed in.
			}\label{fig:RBE}
		\end{figure}
		
		For more clarity, we give the detailed process of relation bias extraction in our proposed Separator Perceiver submodule~(see Sec.~\ref{sec:spg} in main text). As shown in Fig.~\ref{fig:RBE}, given the row/col relations $\mathbf{R}_{row}$ and $\mathbf{R}_{col}$, the corresponding relation masks,  $\mathbf{\Phi}_{row}$ and $\mathbf{\Phi}_{col}$, which have the same sizes with raw table images,  are deduced. In these maps, if two table elements belong to the same row or column relations, the pixels within them would be assigned with the same instance index values, which are denoted by numbers in different colors in the Fig.~\ref{fig:RBE}. As the feature $\mathbf{F}_{enc} = \{D_1, D_2, D_3, D_4\}$ output by deformable encoder is in multiple scales, we directly downsample the relation masks to the size of feature in corresponding scale. Afterward, row and column relation maps are separately embedded with two individual dictionaries via the embedding method~\cite{mikolov2013efficient}. As a result, each of them are encoded to the multi-scale instance index embeddings $\mathbf{M}_{row} = \{M^1_{row}, M^2_{row}, M^3_{row}, M^4_{row}\}$ and  $\mathbf{M}_{col} = \{M^1_{col}, M^2_{col}, M^3_{col}, M^4_{col}\}$, of which the dimensions are 128. After concatenated along channel axis, the embeddings in respective scales are added to the corresponding encoder features. Through this way, the relation information is properly infused into the deformable encoder features.     
		\section{Detailed Descriptions of Datasets}
		\textbf{ICDAR-2013}~\cite{gobel2013icdar} contains 156 tables with ground truth annotated at the word-level, which are collected from natively-digital document. As no training data provided. Similar to~\cite{raja2020table}, we randomly extract 80\% of the data for training and others for testing. Consequently, it is extended to the partial version, \ie, ICDAR-2013-P.
		
		\textbf{ICDAR-2019}~\cite{gao2019icdar} contains 600 training samples and 150 test ones, where all table annotated with cell-level bounding boxes as well as contents.
		
		\textbf{UNLV}~\cite{shahab2010open} is a dataset collected from real world, where the sources ranging from magazines to newspapers.  There are 558 table examples with cell-level box annotations. Considering it is also no training set provided, similar division with ICDAR-2013 dataset is also performed to extend it to the partial version, UNLV-P.
		
		\textbf{TableBank}~\cite{li2020tablebank}  collects 145,000 tables for training and 1,000 ones for testing, where the logical coordinates of cell boxes are provided.
		
		\textbf{PubTabNet}~\cite{end2end1} directly annotates tables as HTML format, which consists of  339,000 training and 114,000 testing samples.
		
		\textbf{SciTSR}~\cite{chi2019complicated} is a large-scale dataset containing 12,000 tables for training and 3,000 testing samples, which are mostly from PDF. Additionally, there are 2,885 and 716 complicated tables are extracted to constitute the SciTSR-COMP~\cite{chi2019complicated}. To evaluate the capacity of TSR methods under more challenging scenes, in work~\cite{liu2022neural} it is further augmented with synthesizing distortion algorithms to simulate distractors brought by capture device, which is called SciTSR-COMP-A. 
		
		\textbf{WTW}~\cite{long2021parsing} is another complex dataset, which contains 10,970 training images and 3,611 testing images collected from wild complex scenes. Specifically, it is mainly composed of the tabular objects with explicit boundaries, with information of table id, tabular cell coordinates and row/column information annotated.

		\section{Details of Loss Function}
		We adopt multi-task loss to supervise the training of our method, as described in the Sec.~\ref{sec:TS} of main text. More concretely, in the separator loss $\mathcal{L}_{sep^{\sim}}$ denoted by Eqn.~\eqref{eq:seploss} of main text, the classification loss $\mathcal{L}_{cls}$ is the standard softmax loss in terms of $\mathbf{u}$, which is defined as: 
		\begin{align}
			&\mathcal{L}_{cls} = -\textup{log}(P(z=c|\mathbf{u})),\\
			&P(z=c|\mathbf{u})=\frac{\exp(S_{c}\mathbf{u})}{\sum_{k}\exp(S_{k} \mathbf{u})}, c \in \{0, 1, 2\},
		\end{align}
		where $\mathbf{u}$ is the output feature of FC layer appended after transformer blocks while $z$ is the predicted class for separator representations. $c \in \{0, 1, 2\}$ corresponds to the explicit or implicit separator, and background otherwise. 
		
		For the structure loss $\mathcal{L}_{struct}$, we introduce the standard cross-entropy loss. Given an ordered source sequence $\mathcal{H}$ sent into the transformer block of Structure Composer, our aim is to minimize the cross entropy loss between ground truth separator indexes and model predictions. The training loss over a single token $\boldsymbol{x}$ is
		\begin{align}
			\mathcal{L}_{struct}(\boldsymbol{y}, \boldsymbol{x})=-\sum_i \sum_e T\left(\boldsymbol{y}_i, e\right) \log p\left(\boldsymbol{y}_i \mid \boldsymbol{y}_{<i}, \boldsymbol{x}\right)
		\end{align}
		where $T\left(\boldsymbol{y}_i, e\right)$ denotes the target label function.
		
		\section{More results on Regularized Table Datasets}
		\begin{table*}[htb!]
			\centering
			\small
			\setlength{\tabcolsep}{0.7mm}
			\scalebox{1} {
				\begin{tabular}{l|l|ccc|l|ccc|l|ccc|l|ccc}
					\toprule[1.5pt]
					&\multicolumn{4}{c|}{\textbf{ICDAR-2013-P}} &\multicolumn{4}{c|}{\textbf{ICDAR-2019}} &\multicolumn{4}{c|}{\textbf{UNLV-P}} &\multicolumn{4}{c}{\textbf{SciTSR}}\\
					\hline
					Method& \tabincell{l}{Train Set} & P & R & F1 & \tabincell{l}{Train Set} & P & R & F1& \tabincell{l}{Train Set} & P & R & F1& \tabincell{l}{Train Set} & P & R & F1\\
					\hline
					{C-CTRNet~\cite{long2021parsing}} & WTW + IC19 & 95.5 &88.3 & 91.7 &WTW& -&-&80.8&-&-&-&-&-&-&-&-\\
					FLAG-Net~\cite{liu2021show} &\tabincell{l}{Sci. + IC13-P} & 97.9& 99.3 &98.6 &\tabincell{l}{Sci. + IC19} &  82.2 & 78.7 &80.4 & \tabincell{l}{Sci. + UNLV-P} &{89.2}&87.3&88.2&Sci.& {99.7} &99.3 &99.5\\
					TSRFormer~\cite{lin2022tsrformer}  & - & - & - & -  &- & - & - & -&-&-&-&-&Sci. & 99.5 & 99.4 & 99.4 \\
					NCGM~\cite{liu2022neural} & \tabincell{l}{Sci. + IC13-P} & 98.4& 99.3   & 98.8 &\tabincell{l}{Sci. + IC19} &84.6 &86.1 &85.3  & \tabincell{l}{Sci. + UNLV-P} & 88.9 & 88.2 & 88.5 &Sci. & 99.7 & 99.6 & 99.6\\
					\hline
					\textbf{GrabTab} & \tabincell{l}{Sci. + IC13-P} & \textbf{98.8} & \textbf{99.5}   & \textbf{99.1} & \tabincell{l}{Sci. + IC19} & \textbf{86.1} & \textbf{86.3} & \textbf{86.2}  &  \tabincell{l}{Sci. + UNLV-P} & \textbf{90.2} & \textbf{90.5} & \textbf{90.3} & Sci.& \textbf{99.8}& \textbf{99.8} & \textbf{99.8} \\
					\bottomrule[1.5pt]
				\end{tabular}
			}
			\caption{Comparison results on ICDAR-2013-P, ICDAR-2019, UNLV-P and SciTSR datasets. ``P'', ``R'' and ``F1'' stand for ``Precision'', ``Recall'' and ``F1-score'' respectively. ``C-CTRNet'' and ``Sci.'' are short for ``Cycle-CenterNet'' and SciTSR.}
			\label{Comparison_results_more}
		\end{table*}	
		
		Tab.~\ref{Comparison_results_more} gives the further results on more regularized table datasets. Although our GrabTab is mainly designed for the complex TSR task, we can still observe the consistent improvements on the regularized table datasets. Especially on the UNLV-P dataset, our method can beat the second best method, NCGM~\cite{liu2022neural}, by 1.8\% F1-score performance. Despite our GrabTab inherits the element relationships from NCGM, thanks to the proposed deliberation mechanism, it can decently exploit multiple table components, where the performance improvements further verify its effectiveness and necessity.	
		
		\section{Visualization of Predicted Results}
		\begin{figure*}[htb!]
			\centering
			\subfigure[Sample results of GrabTab on ICDAR-2013 dataset.] {
				\includegraphics[width=0.98\linewidth]{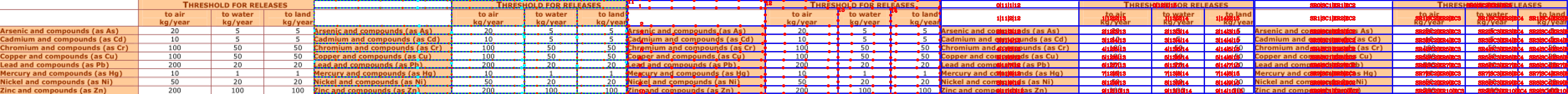}
			}
			\subfigure[Sample results of GrabTab on ICDAR-2019 dataset.] {
				\includegraphics[width=0.98\linewidth]{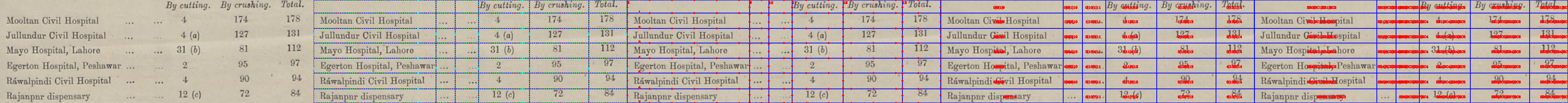}
			}
			\subfigure[Sample results of GrabTab on UNLV dataset.] {
				\includegraphics[width=0.98\linewidth]{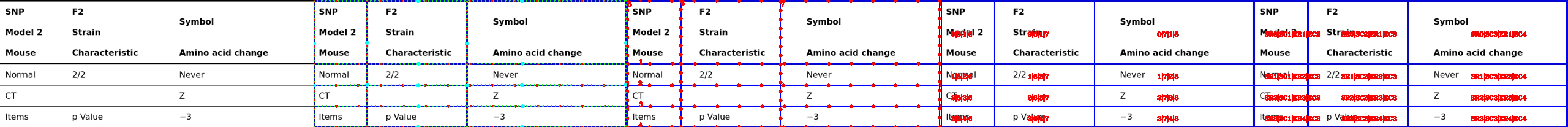}
			}
			\subfigure[Sample results of GrabTab on SciTSR dataset.] {
				\includegraphics[width=0.98\linewidth]{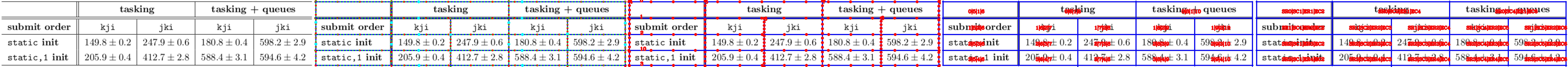}
			}
			\subfigure[Sample results of GrabTab on SciTSR-COMP-A dataset.] {
				\includegraphics[width=0.98\linewidth]{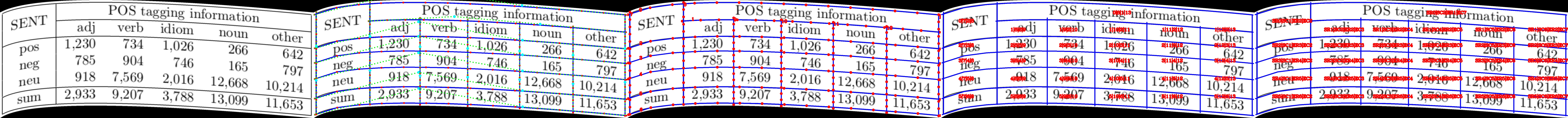}
			}
			\subfigure[Sample results of GrabTab on WTW dataset.] {
				\includegraphics[width=0.98\linewidth,height=3cm]{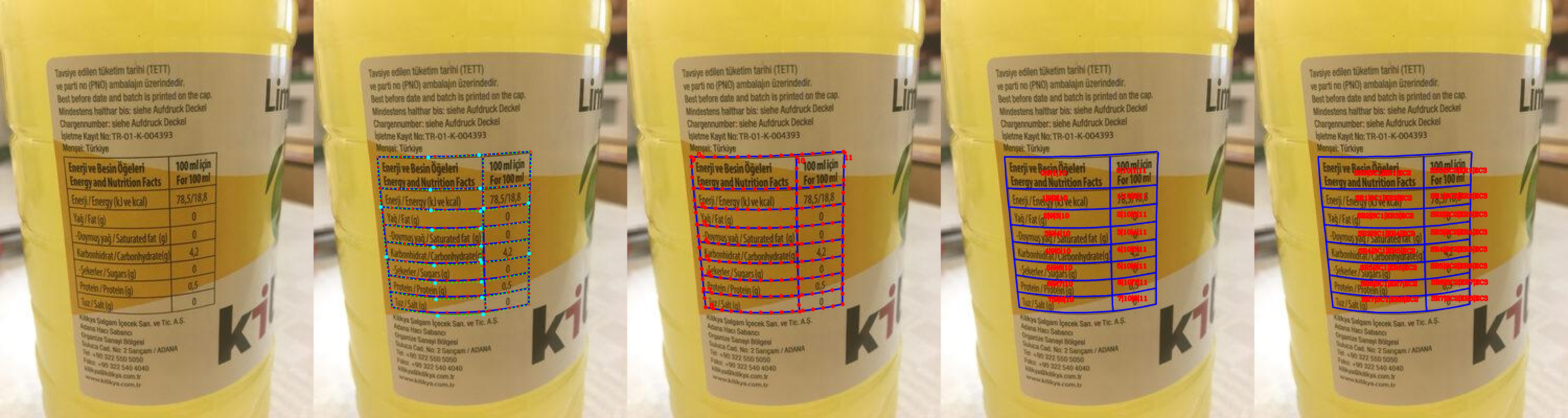}
			}
			\subfigure[Sample results of GrabTab on WTW dataset.] {
				\includegraphics[width=0.98\linewidth]{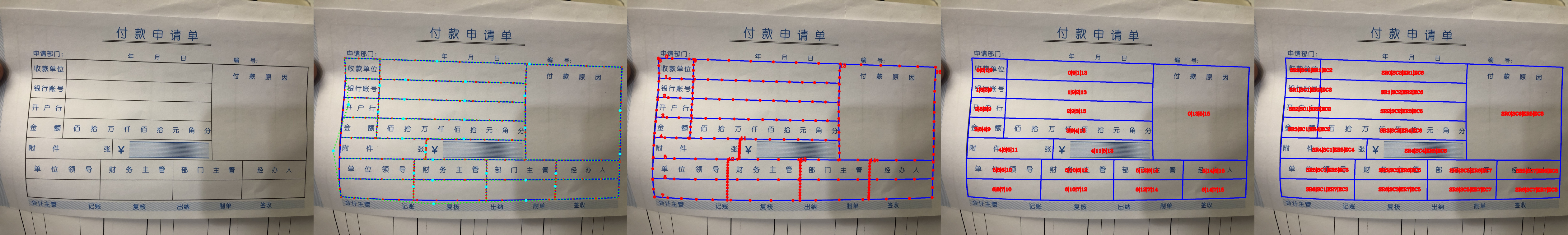}
			}
			\subfigure[Sample results of GrabTab on TableBank dataset.] {
				\includegraphics[width=0.98\linewidth]{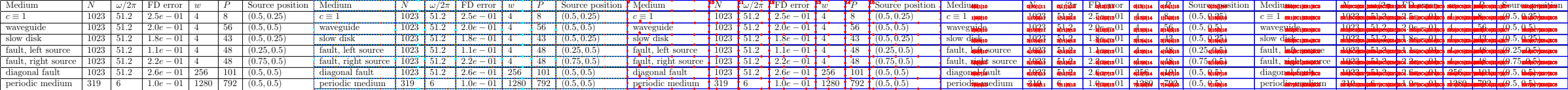}
			}
			\subfigure[Sample results of GrabTab on PubTabNet dataset.] {
				\includegraphics[width=0.98\linewidth]{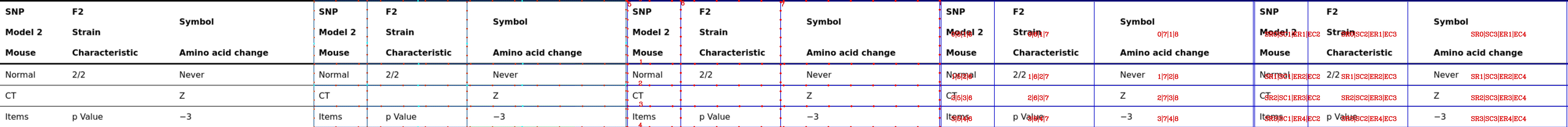}
			}
			\caption{Qualitative results on ICDAR-2013, ICDAR-2019, UNLV, SciTSR, SciTSR-COMP-A, WTW, TableBank and PubTabNet datasets. 1st column: raw images; 2nd column: predicted sample points and fitted separators output by Components Correlator; 3rd column: predicted separators and corresponding order indexes; 4th column: results predicted by Structure Composer; 5th column: Final table structure results.}
			\label{fig:more_vis_result} 
		\end{figure*}
		
		Fig.~\ref{fig:more_vis_result} illustrates more visualizations of predicted results on different datasets. Taking the results on WTW~\cite{long2021parsing}~(Fig.~\ref{fig:more_vis_result}(f)) for example, the first column is the raw table image while the second column shows a set of predicted sample points~(in red color) representing separators. In addition, also in the second column, the cyan color points are the control points of cubic Bézier curve after applying least squares fitting algorithm and blue solid lines indicate the actual separators generated from control points. Moreover, in the third column, the numbers aside the separators represent their orders which correspond to the order output by ``Order Regularization'' submodules in Structure Composer~(SC). In the fourth column, the results predicted by SC are visualized. On each cell, the quadruple numbers in the format of ``$\alpha_{top}| \alpha_{left}| \alpha_{bottom}| \alpha_{right}$'' are the anticlockwise arranged indexes of separators wrapping a cell. As these quadruplets are predicted in sequential format, which naturally contains order information, they can be feasibly converted to the standard table structure format in many existing works, such as NCGM~\cite{liu2022neural}, which is shown in the last column of Fig.~\ref{fig:more_vis_result}(f). Among, ``SR'' and ``SC'' are short for ``start row separator'' and ``start column separator'' while ``ER'' and ``EC'' denote ``end row separator'' and ``end column separator'' respectively. The numbers appended after them are the corresponding indexes.

		\section{More Analysis on GrabTab}
		\subsection{Decoupled Components Correlation}
		\begin{table}[htb]
			\setlength{\tabcolsep}{4mm}
			\centering
			\begin{tabular}{l|ccc}
				\toprule[1.5pt]
				Method	& P & R& F1 \\
				\hline
				{GrabTab-Coupled} & {93.9} & {93.3}  & {93.6} \\
				\textbf{GrabTab} &\textbf{94.3} & \textbf{93.8}  & \textbf{94.0} \\
				\bottomrule[1.5pt]
			\end{tabular}  
			\caption{Performance comparison between GrabTabs with decoupled and coupled Components Correlator on SciTSR-COMP-A  dataset.}
			\label{ablation-cc}
		\end{table}
		
		As one can observe from Tab.~\ref{ablation-cc}, if we modify the Components Correlator into the coupled version, where the both components are simply concatenated and only one stream holistic transformer blocks are preserved, the performance witnesses obvious drop. This phenomenon demonstrates the indispensability of decoupled structure, which is able to avoid the mutual effect between candidate components when they providing informative clues for constructing table structure.     
		\subsection{Improvements on SOTA Methods}
		\begin{table}[htb]
			\centering
			\begin{tabular}{l|ccc}
				\toprule[1.5pt]
				Method	& P & R& F1 \\
				\hline
				GraphTSR & 74.9 & 76.4 & 75.6  \\
				GraphTSR + GrabTab & {93.3} & {93.5}  & {93.4} \\
				\hline
				DGCNN & 76.3 & 77.1  & 76.7  \\
				DGCNN + GrabTab & {93.8} & {93.2}  & {93.5} \\
				\hline
				\textbf{GrabTab} &\textbf{94.3} & \textbf{93.8}  & \textbf{94.0} \\
				\bottomrule[1.5pt]
			\end{tabular}  
			\caption{Performance comparison of GrabTab under different relation extraction components on SciTSR-COMP-A  dataset.}
			\label{tab:sota_grab}
		\end{table}
		
		Functionally, as our GrabTab focuses on the multiple components collaboration, it can play as a plug-and-play module combining with other methods. To further investigate its facilitation to existing methods, we combine two element relationship-based SOTAs, including GraphTSR~\cite{chi2019complicated} and DGCNN~\cite{wang2019dynamic}, with our GrabTab. Specifically, the candidate components are firstly extracted from them, which are sent to the deliberator in our GrabTab. From Tab.~\ref{tab:sota_grab}, we surprisingly observe the significant improvements~(round 20\%) on the performance of both methods, which could be attributable to the decent compatibility to the complex tables brought by our GrabTab, including more flexible separator representation and efficient components employment.

		\section{Computational Complexity}\label{sec:complexity}
		\begin{table}[htb]
			\centering
			\setlength{\tabcolsep}{1.5mm}
			\begin{tabular}{l|cccc}
				\toprule[1.5pt]
				Method & \#Param & FLOPs & GPU & CPU \\
				\hline
				SPLERGE~\cite{tensmeyer2019deep} & 0.37 & 115.49 & 0.95 & 24.25 \\
				TabStruct-Net~\cite{raja2020table} & 68.63 & 3719.06 & 22.63 & 76.52 \\
				FLAG-Net ~\cite{liu2021show}  & 17.00 & 71.69 & 0.13 & 2.37 \\
				\hline 
				GrabTab & 57.88& 513.65 & 0.82 & 11.54 \\
				\bottomrule[1.5pt]
			\end{tabular}
			\caption{The inference efficiency of different methods. \#Param denotes the number of parameters, while FLOPs are the numbers of FLoating point OPerations.
				The units are million (M) for \#Param, giga (G) for FLOPs, second (s) for GPU time, and second (s) for CPU time. The execution time is computed on one Nvidia Tesla V100 GPU and a 2.4 GHz Intel Xeon E5 CPU.
			}
			\label{tab:param2}
			\centering
		\end{table}
		Tab.~\ref{tab:param2} gives the comparison on the computational complexity of different
		methods, where the model sizes and the inference operations of different models are summarized. Concretely, GPU/CPU inference time is calculated by averaging time consumption on all datasets. 
		As the Structure Composer in our model is essentially a generative model which predicts the final results in a recursive way, the  inference time and FLOPs are inevitably increased. It is worthy noting that, FLAG-Net can achieve the most economic computational consumption, however, it can not well handle the complex table cases. Even if we increase the model capacity to the similar level with our method, the complex TSR problem could not theoretically solved under the FLAG-Net framework. In contrast, the performance of our GrabTab can surpass it by a large margin on complex TSR datasets~(round 5\% on WTW and 12\% on SciTSR-COMP-A) as illustrated by Tab.~\ref{Comparison_results} of main text.

		\section{Limitations and Future Works}
		As introduced throughout this paper, our main purpose is to explore the positive effects of multiple components collaboration through deliberation mechanism, rather than the extraction in terms of exact specific component. Therefore, in the current version, we simply extract the table elements and and their relationships in off-line manner. Alternatively, a more potential way is to repurpose the candidate components extraction into an on-line one, with the model weight updated together with the subsequent deliberator. Besides, as aforementioned in Sec.~\ref{sec:complexity}, the increase on the computational complexity is another issue to be solved we leave in our future work.
	\end{appendices}
	{\small
		\bibliographystyle{ieee_fullname}
		\bibliography{grabtab.bib}
	}

\end{document}